\documentclass[10pt,twocolumn,letterpaper]{article}

\usepackage{cvpr}              
\usepackage{multicol}
\usepackage{multirow}
\usepackage{float}
\usepackage{placeins}
\usepackage{dblfloatfix} 

\usepackage[dvipsnames]{xcolor}

\definecolor{cvprblue}{rgb}{0.21,0.49,0.74}
\usepackage[pagebackref,breaklinks,colorlinks,citecolor=cvprblue]{hyperref}

\def\paperID{160} 
\def\confName{3DV\xspace}
\def\confYear{2024\xspace}

\title{MuVieCAST: Multi-View Consistent Artistic Style Transfer}

\author{Nail Ibrahimli, Julian F. P. Kooij, Liangliang Nan\\
Delft University of Technology\\
Julianalaan 134, Delft 2628BL, The Netherlands\\
{\tt\small \{n.ibrahimli, j.f.p.kooij, liangliang.nan\}@tudelft.nl}\\
}

\begin{document}
\maketitle
\begin{abstract}
We introduce MuVieCAST, a modular multi-view consistent style transfer network architecture that enables consistent style transfer between multiple viewpoints of the same scene. This network architecture supports both sparse and dense views, making it versatile enough to handle a wide range of multi-view image datasets. The approach consists of three modules that perform specific tasks related to style transfer, namely content preservation, image transformation, and multi-view consistency enforcement. We extensively evaluate our approach across multiple application domains including depth-map-based point cloud fusion, mesh reconstruction, and novel-view synthesis. 
Our experiments reveal that the proposed framework achieves an exceptional generation of stylized images, exhibiting consistent outcomes across perspectives. A user study focusing on novel-view synthesis further confirms these results, with approximately 68\% of cases participants expressing a  preference for our generated outputs compared to the recent state-of-the-art method. Our modular framework is extensible and can easily be integrated with various backbone architectures, making it a flexible solution for multi-view style transfer.
More results are demonstrated on our project page: \href{https://muviecast.github.io/}{muviecast.github.io}

\end{abstract}    
\section{Introduction}
\label{sec:intro}

Style transfer has gained popularity in computer vision for its ability to create unique and visually appealing images~\cite{gatys2016image,chen2016fast,li2016combining,risser2017stable,gu2018arbitrary,kolkin2019style,Huang_2017_ICCV,liao2017visual,Johnson2016Perceptual,WCT-NIPS-2017,ulyanov2016instance,zhang2017multistyle}. 
Conventional style transfer techniques ~\cite{gatys2016image,Huang_2017_ICCV,Johnson2016Perceptual} involve transforming an input image to mimic the style of a reference image. 
3D stylization methods instead transfer styles between 3D scene representations~\cite{KPLD21,cao2020psnet,Huang22StylizedNeRF,huang2021learning,zhang2022arf,hollein2021stylemesh}. These methods typically operate on input formats such as point clouds~\cite{KPLD21,cao2020psnet,huang2021learning}, meshes~\cite{hollein2021stylemesh}, or neural radiance fields~\cite{zhang2022arf,Huang22StylizedNeRF} to generate style transfers for a single 3D scene representation.

In this work, we address multi-view style transfer, i.e.~the challenge of maintaining consistency across multiple views of a static scene or object.
We aim for a universal solution that works well in diverse 3D scene representation domains, removes the need for groundtruth data or precomputed 3D information, and can adapt to different datasets and view coverages~\cite{schoenberger2016mvs}.

More concretely, we seek a function, $f(\mathbf{I}, \mathbf{C}, \mathbf{S})$, that takes a set of calibrated multi-view images of the scene, $\mathbf{I} = {I_1, I_2, ..., I_N}$, with corresponding camera parameters, $\mathbf{C} = {C_1, C_2, ..., C_N}$, and a style image, $\mathbf{S}$, and outputs styled images that maintain both photometric and style consistency across all views.  In practice, we can use Structure from Motion (SfM)~\cite{schoenberger2016sfm,moulon2013global,Moulon2012} to estimate camera parameters from input images when they are not available. Formally, we define this function as $f(\mathbf{I}, \mathbf{C}, \mathbf{S}) = \mathbf{I}'$, where $f$ transforms each input image $I_i$ to $I'_i$ to adopt the style of image $\mathbf{S}$.
The function should capture the knowledge of both the 3D geometry and 2D image structure, while retaining the input image's content and applying the desired style, ultimately generating multi-view consistent images.

To address these challenges, we propose MuVieCAST, a modular framework for consistent, flexible, and fast style transfer between multiple views of the  scene.  MuVieCAST comprises multiple modules to execute various tasks including feature extraction, style transfer, and consistency enforcement. Its modular design enables integration with different computer vision backbones.

It offers unconstrained and pretrained options for style transfer, giving flexibility in choosing styles. 
Furthermore, we present multiple backbone options for customized network architectures, such as geometry learning~\cite{gu_2020_cascademvsnet,wang2020patchmatchnet}, feature extraction~\cite{simonyan2014very}, and image transformation~\cite{ronneberger2015u,Huang_2017_ICCV}.

In summary, our modular multi-view consistent style transfer framework offers a fast, versatile, and robust solution that accommodates both sparse and dense views, provides flexibility in terms of arbitrary and pretrained options for style transfer, and provides multiple backbone options, making it suitable for a diverse range of tasks and applications, such as novel-view synthesis, point cloud reconstruction, and mesh reconstruction.

\section{Related work}
\label{sec:related}

Our work is related to neural style transfer for images, videos, and 3D scenes. We discuss each in turn.

\subsection{Single-view neural style transfer} 

Neural style transfer~\cite{gatys2016image,Johnson2016Perceptual,li2017demystifying} is a technique that utilizes pretrained convolutional neural networks to extract features from a content image and a style image. 
The goal is to transfer the style of the style image onto the content image by optimizing for loss functions. 
There are two types of neural style transfer methods: optimization-based and feed-forward-based approaches.

Optimization-based methods involve iterative optimization to minimize content and style losses.  
The popular approach involves calculating the Gram matrix~\cite{gatys2016image}, which captures the correlations between feature maps, to compute the style loss. 
Researchers have explored alternative style loss formulations to enhance semantic consistency and high-frequency style details. Several methods~\cite{chen2016fast,li2016combining, kolkin2022neural, liao2017visual} use nearest neighbor search and minimize distances between features extracted from corresponding content and style patches in a coarse-to-fine manner. 

The feed-forward-based methods use neural networks to transfer the style information of the style image to the input image in a single forward pass.
Though the naive feed-forward approach is fast~\cite{Johnson2016Perceptual}, it requires training on specific styles and may not generalize well to arbitrary styles. 
To address this issue, Huang et. al.~\cite{Huang_2017_ICCV} use first-order statistics to encode the style information and transform the image style via AdaIN layers. 
This method matches the mean and variance of intermediate features between content and style images. 
This allows for more flexible style transfer by adjusting the scaling and shifting of the content features based on the statistics of the style features. 
The WCT~\cite{WCT-NIPS-2017} approach further explores the covariance instead of the variance, and it also uses whitening and coloring transformation to match the second-order statistics of the input image to those of the reference image. 
Additionally, the LST~\cite{li2018learning} scheme leverages convolutional neural networks to reduce the computational cost of solving the transformation matrix in the WCT~\cite{WCT-NIPS-2017} method for real-time universal style transfer.
Several works ~\cite{liu2017depth,ioannou2022depth} utilize a monocular depth prediction network to integrate depth preservation to~\cite{Johnson2016Perceptual} as an additional loss function.

\subsection{Video stylization}
Video stylization aims to transfer the style of a style image to a sequence of video frames. To solve the problem of flickering that arises with image stylization methods, many techniques~\cite{Chen_2017_ICCV,Chen2020OpticalFD,Gao2018ReCoNetRC,Gupta_2017_ICCV,Huang_2017_CVPR}  use optical flow constraints to train autoencoders networks that can transfer a specific style to videos. Recent developments have allowed video style transfer to be performed for arbitrary styles~\cite{deng2020arbitrary,gao2020,Compound2020}. However, as shown by Huang et. al.~\cite{huang2021learning}, applying these methods for 3D scene stylization can result in undesirable outcomes, including blurry or inconsistent images across different novel views. In light of this, our focus in this work is to enforce consistency across multiple views.

\subsection{3D style transfer} 
The goal of 3D style transfer is to modify the appearance of a 3D scene to match the style of a desired image when rendered from different viewpoints. 
Previous methods represent real-world scenes as point clouds, triangle meshes, or radiance fields, which are then fed as input for neural networks to produce stylized renderings. For instance, some methods~\cite{huang2021learning,mu20223d} utilize 3D point clouds that are first processed with the desired style and then passed through a convolutional renderer. Stylemesh ~\cite{hollein2021stylemesh} applies style transfer on mesh reconstructions of indoor scenes. Recent methods~\cite{zhang2022arf, Huang22StylizedNeRF} stylize radiance fields~\cite{mildenhall2020nerf,yu2021plenoctrees,Chen2022ECCV} which are effective for novel view synthesis of real-world scenes captured at high density. 
However, common NeRF backbones used in style transfer~\cite{mildenhall2020nerf,yu2021plenoctrees,Chen2022ECCV,kar2017learning, Huang22StylizedNeRF}  require dense view coverage to reconstruct accurate radiance fields.  Few-shot learning NeRF methods~\cite{truong2023sparf,niemeyer2022regnerf} are known to have limitations in handling
large baseline shifts, which impact the quality of reconstructed radiance fields.

Contrary to existing methods, MuVieCAST is flexible with the number of views (dense and sparse) and shows no need for precomputing a 3D representation.
It directly learns to generate stylized views and produce high-quality stylizations within a short training time. The generated outputs are sufficiently consistent, making them suitable for various downstream applications such as depth estimation, point cloud and mesh reconstruction, and novel-view synthesis.
\section{Method}
\label{sec:method}

\begin{figure}[ht]
	\centering
	\includegraphics[width=0.98\columnwidth]{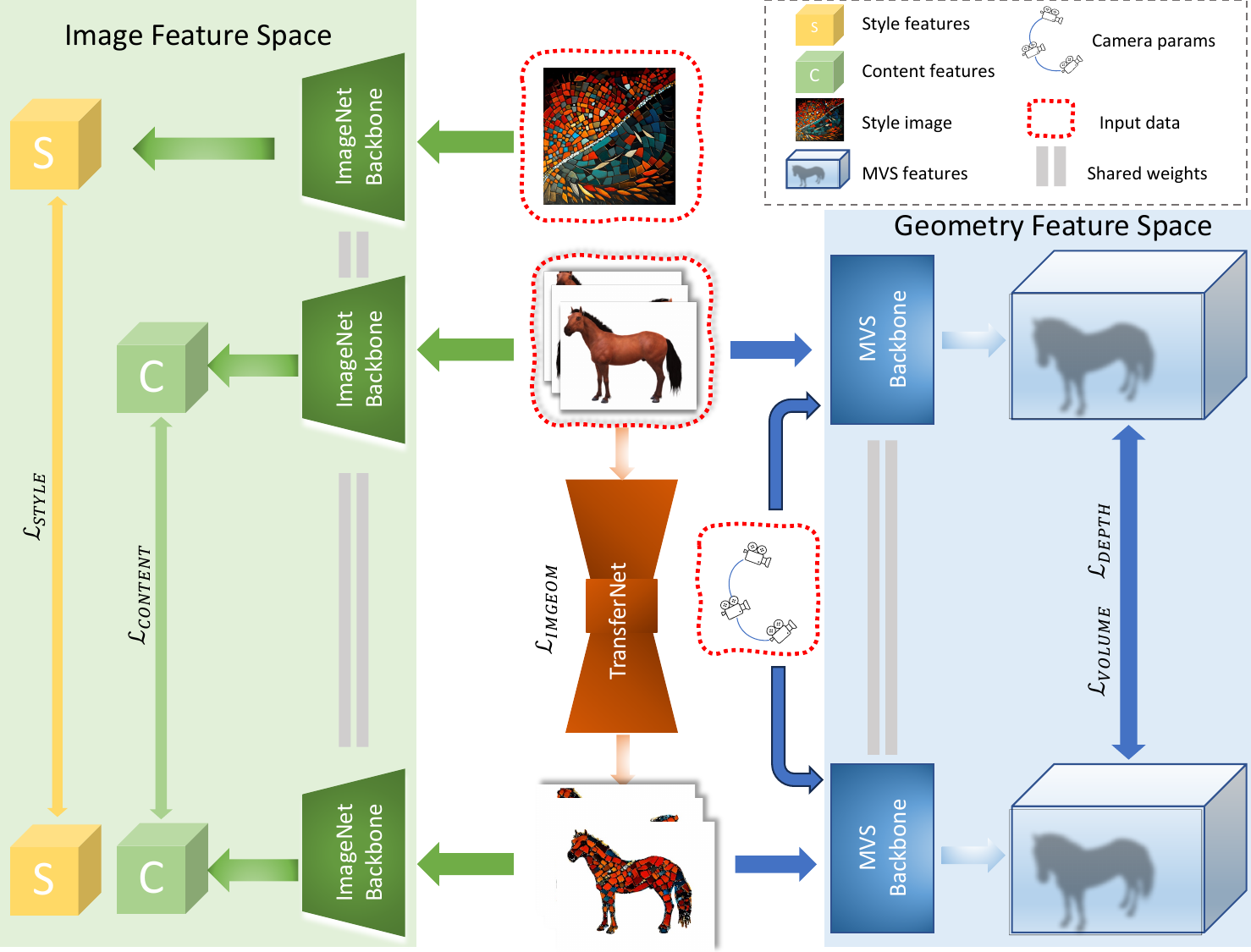}
    \caption{Our proposed network structure (MuVieCAST) has three main components: \textbf{Content-style feature extraction} (green) operates on the image feature space for preserving the content and style. \textbf{TransferNet} performs image transformation. \textbf{Geometry learning module} (blue) operates on the geometry feature space for preserving the geometry.}

	\label{fig:network_architecture}
\end{figure}

In this section, we present our novel approach for multi-view consistent style transfer, which transfers the style of an artistic image to a set of target images of the same scene or object while maintaining consistency across multiple views. 
As shown in Fig~\ref{fig:network_architecture}, our method comprises three main components: style transfer network (TransferNet), content-style feature extraction,  and a multi-view stereo (MVS) based geometry learning module.

First, we use TransferNet to transform the input RGB space image to the style space. 
Then, we extract feature representations from both the input and transformed images using a pretrained deep neural network. These features guide the style transfer network to generate images with similar artistic features while preserving the content.

To ensure consistency across multiple viewpoints, we utilize an MVS-based geometry learning module that leverages the geometric information provided by multiple views. 
This module uses input and transformed images as well as camera poses to estimate the 3D geometry of the scene and enforce style consistency across different views.

 These components work together to produce high-quality and consistent style transfer results, even in complex multi-view scenarios.
 In the next subsection, we detail each component, including the content-style feature extraction, the style transfer network, and the MVS-based geometry learning module.

 \subsection{Modules}

 \paragraph{TransferNet}

Our approach employs a TransferNet module for generic style transfer. In contrast to prior multi-view approaches~\cite{Huang22StylizedNeRF,zhang2022arf}, our method utilizing image-geometry loss guidance aims to enhance the texture while preserving the primary gradient structure by considering the first and second-order spatial derivatives of the original input views. This approach also assists in maintaining the finer details of the individual scene image following the stylization. Additionally, we incorporate 3D geometry loss guidance to ensure that the resulting image conforms to the overall scene geometry captured from multiple views.
 
 \paragraph{Content-style feature extraction}
We utilize content style feature reconstruction, which extracts content and style features from pretrained Imagenet~\cite{deng2009imagenet}  backbones~\cite{simonyan2014very}.
Pretrained weights from Imagenet are used to extract the features, and the resulting content style features guide the TransferNet to ensure that the content of the input image is preserved while the style of the artistic image is transferred onto it. 
It is important to note that instead of optimizing the weights of these backbones, we use them solely to extract features and guide the TransferNet.
This approach allows us to focus the fast network optimization on the TransferNet, 
rather than learning content extraction from scratch for a longer period.
By using pre-existing knowledge from the pretrained models, the TransferNet not only leads to higher quality style transfer results but also significantly reduces the computational cost of the optimization process.

 \paragraph{Geometry learning module}

We choose to use multi-view stereo-based approaches as our geometry learning backbones for several reasons. 
First, these methods are generally applicable and can be used without further tuning.
Additionally, we can leverage the pretrained modules and focus on optimizing the weights of TransferNet, which makes them particularly attractive for our multi-view consistent style transfer framework. 
Unlike common neural rendering or neural radiance fields approaches~\cite{yariv2020multiview,yariv2021volume,wang2021neus,mildenhall2020nerf,yu2021plenoctrees, Huang22StylizedNeRF,zhang2022arf} used for style transfer, these networks can also robustly work with sparse views, making them more applicable in a wide range of scenarios.
We discuss an ablation study involving sparse views within our supplementary materials.

Our framework supports multi-view stereo backbones \cite{gu_2020_cascademvsnet,wang2020patchmatchnet} which employ a feature pyramid network~\cite{lin2017feature} to extract multiple levels of features and guide the stereo matching process. The coarse-to-fine approach involves processing the images at different scales, starting with a low resolution and gradually increasing it. This makes our backbones more memory-efficient and able to handle larger input sizes~\cite{yang_2020_cvpr,gu_2020_cascademvsnet,wang2020patchmatchnet}.

\subsection{Loss}

Our loss function is a weighted combination of loss terms.  These loss terms aim to achieve multi-view consistent style transfer, preserving the primary image and geometric structures of the original views while transforming them into the generated stylized views.

\paragraph{Content-style loss}
To achieve content-aware style transfer, we use a combination of content loss and style loss.
The content loss measures the difference between the feature representations of the target image and the reference image, which serves as the source of the content information.
Let $F_t$ and $F_r$ be the feature maps of the target image and input reference image,  respectively and $\Omega_F$ be the number of elements in feature maps.
The content loss is then defined as follows,
\begin{equation}
\mathcal{L_\text{content}} = \frac{1}{\Omega_F}\sum \lVert F_{t} - F_{r}\rVert^2_2  \; .
\end{equation}

We provide two options for the style loss $\mathcal{L_\text{style}}$. The first option is to measure the difference between the Gram matrices ~\cite{gatys2016image} of the feature maps of the target image and the style image, which contains the texture and color information. Let $G_t$ and $G_s$ be the Gram matrices of the feature maps of the target image and style image, respectively, and $\Omega_G$ be the number of elements in the Gram matrix, such that the style loss is defined as
\begin{equation}
\mathcal{L_\text{style}} = \frac{1}{\Omega_G}\sum \lVert G_{t} - G_{s}\rVert^2_2 \; .
\end{equation}
Our second option is inspired by the work of Li et al.~\cite{li2017demystifying} which explores a style loss based on instance normalization statistics.
This style loss is defined as
\begin{multline}
    \mathcal{L_\text{style}} =\sum_{i=1}^{L}\lVert \mu(F_{i,t}) - \mu(F_{i,s}) \rVert^2_{2}  \quad + \\
\sum_{i=1}^{L}\lVert \sigma(F_{i,t}) - \sigma(F_{i,s}) \rVert^2_{2} \; ,
\label{equ:style}
\end{multline}
where $F_{i,t}$ and $F_{i,s}$ are the feature maps of the generated image and style image at layer $i$, respectively, $L$ is the total number of layers used, $\mu$ is the mean, and $\sigma$ is the standard deviation. The style loss is calculated by computing the L2 norm between the instance normalization statistics of the feature maps of the generated and style images.

\paragraph{Image-geometry loss}

To preserve the primary geometric structures of the input image in the generated image, we introduce an image-geometry loss which consists of three terms: Sobel filter loss, Laplacian operator loss, and differentiable Canny edge detector loss.


These three components capture the geometric characteristics of both the input and generated images. The overall image-geometry loss is defined as the weighted sum of the Sobel filter loss ($\mathcal{L}_\text{Sobel}$), Laplacian operator loss ($\mathcal{L}_\text{Laplace}$), and Canny edge detector loss ($\mathcal{L}_\text{Canny}$), which is expressed as follows:

\begin{align}
\mathcal{L}_\text{Sobel} &= \frac{1}{\Omega_I}\sum \text{smoothL1}(\nabla(I_r),\nabla(I_t)) \\
\mathcal{L}_\text{Laplace} &= \frac{1}{\Omega_I}\sum \text{smoothL1}(\Delta(I_r),\Delta(I_t)) \\
\mathcal{L}_\text{Canny} &= \frac{1}{\Omega_I}\sum \text{smoothL1}(C(I_r),C(I_t)) \\
\mathcal{L}_\text{imgeom} &= \lambda_{\nabla} \mathcal{L}_\text{Sobel} + \lambda_{\Delta} \mathcal{L}_\text{Laplace} + \lambda_\text{C} \mathcal{L}_\text{Canny}\;,
\end{align}

where $\nabla$ is the Sobel filter operator, $\Delta$ is the Laplacian operator, and $C$ is the differentiable Canny edge detector. $I_r$ and $I_t$ correspond to the input and generated images, respectively, and $\Omega_I$ represents the total number of pixels in the images. The hyperparameters $\lambda_\nabla$, $\lambda_\Delta$, and $\lambda_\text{C}$ control the relative influence of each component within the overall image-geometry loss.

The smooth L1 loss function is employed to compute the discrepancy between the responses of the input and generated images for each of the three components.  
The inclusion of these geometric loss components in the overall loss function enables effective regularization of the neural network, preventing excessive divergence from the main geometric features of the input image.

\paragraph{3D geometry loss}

To enforce geometric consistency between the input and output images in our multi-view consistent style transfer framework, we incorporate two types of losses: volumetric loss and depth loss. Both losses use the smooth L1 loss function to calculate the differences between the corresponding elements of the input and target images.

We apply a coarse-to-fine strategy for computing these losses. Specifically, we calculate the losses at multiple stages, with a weight of $2^{3-l}$ at each stage $l$, where $l=0$ is the finest stage. This allows the network to gradually refine the output image.

For the volumetric loss, we use an estimated probability volume for each depth and calculate the distance between the probability volumes of the input and target images. For the depth loss, we first estimate the depth map of the stylized target image and the input images using multi-view stereo backbones. We then calculate the smooth L1 loss between the estimated depth map of the input images and the generated target images. These losses are computed at each stage of the network as

\begin{align}
\mathcal{L_\text{volume}} &= \sum_l\frac{ 2^{3-l}}{\Omega_{V_{(l)}}} \cdot \sum \text{smoothL1} (V_{r(l)} , V_{t(l)} ) \\\
\mathcal{L_\text{depth}} &= \sum_l\frac{ 2^{3-l}}{\Omega_{D_{(l)}}} \cdot \sum \text{smoothL1} (D_{r(l)} , D_{t(l)} ) \; ,
\end{align}
where $\Omega_{V_{(l)}}$ is the total number of voxels in the probability volume at stage $l$. $V_{r(l)}$ and $V_{t(l)}$ are the probability volumes of the depth of the input and target images respectively, at stage $l$. $\Omega_{D_{(l)}}$ is the total number of pixels in the depth map at stage $l$.  $D_{t(l)}$ is the depth values of the stylized image at stage $l$, and $D_{r(l)}$ is the estimated depth values.

\paragraph{Total loss}
Total loss function is a weighted combination of loss terms for the content-style ($\mathcal{L_\text{content}}$ and $\mathcal{L_\text{style}}$), for the image-geometry ($\mathcal{L_\text{imgeom}}$), and for the 3D geometry ($\mathcal{L_\text{volume}}$ and $\mathcal{L_\text{depth}}$), i.e.
\begin{multline}
\mathcal{L_\text{total}} =  \lambda_\text{content} \mathcal{L_\text{content}} + \lambda_\text{style} \mathcal{L_\text{style}} \quad +\\
 \lambda_\text{imgeom} \mathcal{L_\text{imgeom}} +\lambda_\text{volume} \mathcal{L_\text{volume}} + \lambda_\text{depth} \mathcal{L_\text{depth}} \; .
\end{multline}
\section{Experiments}
\label{sec:exp}

In this section, we describe our experiment setup and compare different backbones for our approach. We demonstrate our framework’s multi-view consistency on three applications: depth-map-based point cloud fusion, mesh reconstruction, and novel-view synthesis.

\subsection{Experiment setup}
To demonstrate the modularity of our approach, we perform experiments using several backbones.
Specifically, 
we utilized pretrained VGG16 and VGG19 modules, which were trained on the ImageNet dataset~\cite{deng2009imagenet}, as well as CasMVSNet and PatchmatchNet geometric multi-view stereo backbones pretrained on DTU dataset~\cite{aanaes2016_dtu}. Additionally, we used AdaIN and UNet as style transfer backbones, which were pretrained on the MS COCO dataset~\cite{lin2014microsoft}.
Table~\ref{tab:backbones} summarizes these backbones tested in our experiments.
To show different network configurations, we limit our discussion to four configurations, which are presented in Table~\ref{tab:configurations}.

\begin{table}[h]
\centering
\begin{tabular}{|c|c|c|c|}
\hline
\textbf{Modules} & \textbf{Options} & \textbf{Pretrained} & \textbf{Trainable}\\
\hline
\multirow{2}{*}{\shortstack{Image\\ learning}   }  & VGG16 & ImageNet & no   \\
\cline{2-4}
   & VGG19 & ImageNet & no   \\
\hline
\multirow{2}{*}{\shortstack{Geometry\\ learning}}  & CasMVSNet & DTU & no   \\
\cline{2-4}
  & PatchMatchNet & DTU & no   \\
\hline
\multirow{2}{*}{TransferNet} & UNet & MS COCO & yes   \\
\cline{2-4}
& AdaIN & MS COCO & yes   \\
\hline
\end{tabular}
\caption{Backbones used in our experiments.}
\label{tab:backbones}
\end{table}
 
\textbf{Geometry learning module}. 
PatchmatchNet~\cite{wang2020patchmatchnet} is a patchmatch stereo-based approach~\cite{galliani_2015_gipuma}, where a similarity score is computed between the extracted features of each image in a coarse-to-fine manner to generate a depth map.
CasMVSNet~\cite{gu_2020_cascademvsnet} uses cascaded cost volume regularization to construct a cost volume~\cite{yao_2018_mvsnet,yang_2020_cvpr,collins1996space} and estimate the depth between the images. To further reduce GPU memory demands, we employ groupwise correlation~\cite{xu2020learning_inverse,guo2019group} and in-place activated batch normalization~\cite{rotabulo2017place} within CasMVSNet.

\textbf{TransferNet}. 
The UNet backbone is a fully convolutional neural network that is widely used in image segmentation tasks~\cite{ronneberger2015u}. 
In our framework, we pretrain UNet for each specific style using the MS COCO object detection dataset ~\cite{lin2014microsoft}. 
The pretraining is performed to learn the style-specific weights of the network, which are then tuned for style transfer. 
In the process of multi-view consistent style transfer, the optimization is initialized with these pretrained weights.
The other backbone, AdaIN (Adaptive Instance Normalization), is a style transfer technique~\cite{Huang_2017_ICCV,Huang22StylizedNeRF} that does not require style-specific pretraining. 
Instead, it uses a single encoder network to extract content and style information from both the content and style images. 
The encoder network maps the style image to the feature space, and the content image features are transformed to this feature space using adaptive instance normalization. 
Finally, the transformed content features are mapped back to the image space by the decoder to generate the stylized output. This decoder is also initialized with pretrained weights on the MS COCO dataset~\cite{lin2014microsoft}.


In the supplementary material, we expand on our findings about using UNet's pretrained module for quick adaptation to new styles and images. By using weights learned from MS COCO object detection, we can swiftly apply UNet to new style images for given inputs. Pretraining captures a general understanding of style features and textures that can be reused, significantly reducing the time and resources needed for training new models for each style.

While both network configurations yielded visually appealing results, our experiments showed that UNet-based configurations demonstrated greater accuracy and consistency in maintaining geometry. 
For each experiment on point cloud and mesh reconstruction, we trained our model for 10 epochs, which took less than 5 minutes to complete on dual RTX2080 Ti.

\begin{table*}[h]
\centering
\label{tab:configs}
\begin{tabular}{|c|c|c|c|c|c|c|}
\hline
\textbf{Naming} & \textbf{Geometry} & \textbf{ImageNet} & \textbf{TransferNet} & \textbf{Style loss} & \textbf{Total params} & \textbf{Trainable params}  \\
\hline
CasMVSNet\_UNet  & CasMVSNet & VGG16 & UNet & Gram & 10.2 M & 1.7 M   \\
\hline
CasMVSNet\_AdaIN & CasMVSNet & VGG19 & AdaIN & IN statistics & 7.9 M & 3.5 M \\
\hline
PatchMatchNet\_UNet  & PatchMatchNet & VGG16 & UNet & Gram & 9.5 M & 1.7 M\\
\hline
PatchMatchNet\_AdaIN & PatchMatchNet & VGG19 & AdaIN & IN statistics & 
7.2 M & 3.5 M\\
\hline
\end{tabular}
\caption{Different network configurations tested in the experiments.}
\label{tab:configurations}
\end{table*}

\subsection{Point cloud reconstruction}
 \begin{figure}[]
	\centering
	\includegraphics[width=0.98\columnwidth]{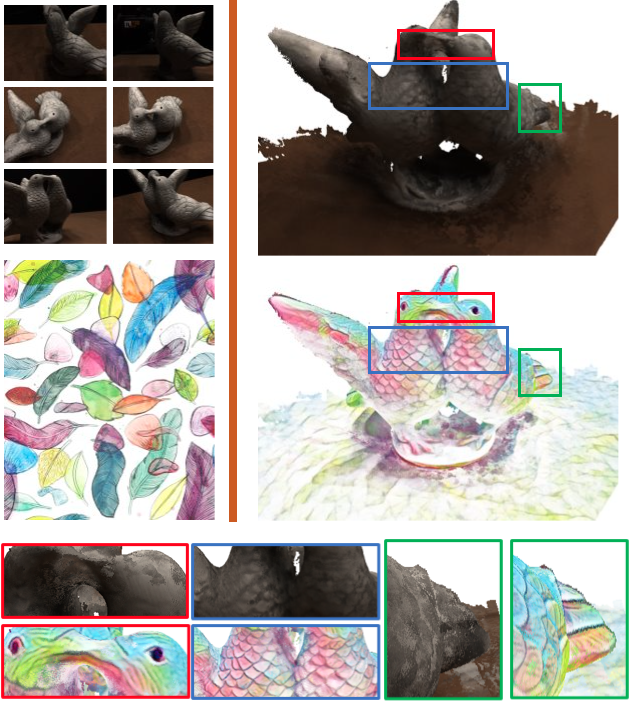}
	\caption{The bird point cloud reconstructed using depth map fusion from the 49 input images \cite{aanaes2016_dtu} and a style image. This experiment demonstrates the stylization capability of our network  (CasMVSNet\_UNet) to reveal finer details in textured areas, even in shadowed regions. Both point clouds have identical geometry.}
	\label{fig:pointcloud2}
\end{figure}

We found that our network is resilient to lighting variations in the input and attentive to texture details, even in shadowed regions. 
This capability becomes particularly evident when the network is trained using style images characterized by significant color variations.
Figure \ref{fig:pointcloud2} illustrates the point cloud reconstruction generated from the input views using depth map fusion \cite{yao_2018_mvsnet, ibrahimli2023ddl}, which uses learning-based multi-view stereo matching \cite{gu_2020_cascademvsnet}. 
Both point clouds contain the same set of points.
Since the input images were captured under different lighting conditions, the coloring of the point cloud is blurred and finer details like the birds' eyes and body textures are not properly revealed. However, our network colors the point cloud with better detail. 
 \begin{figure}[]
	\centering
	\includegraphics[width=0.96\columnwidth]{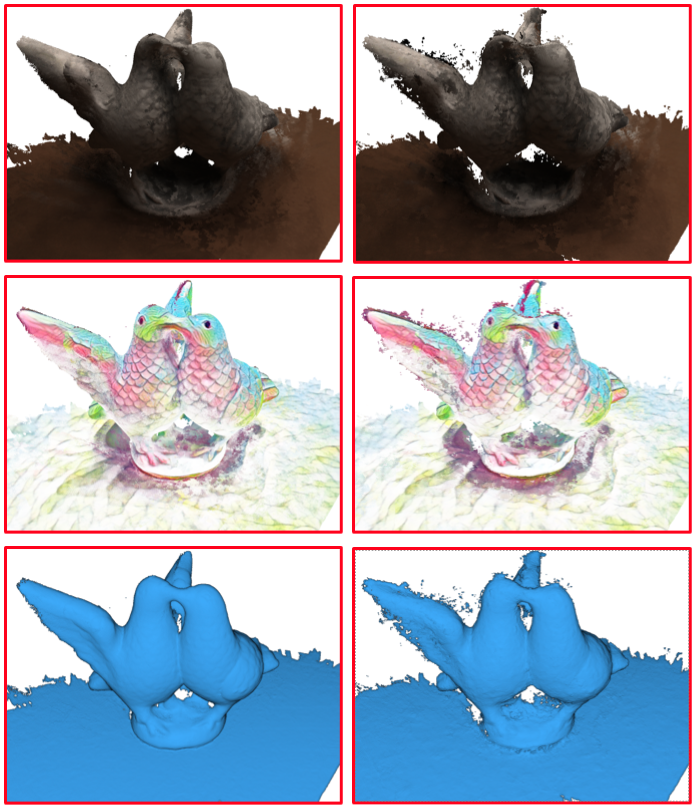}
	\caption{Comparison of the point clouds reconstructed from the original input images (left column) and stylized images (right column). Top row:  colored with original inputs. Middle row: colored with stylized colors. Bottom row: uniform coloring.
    The reconstructed point cloud from stylized views (CasMVSNet\_UNet) closely resembles that from the original input, confirming that the stylization is multi-view consistent.
 }
	\label{fig:pointcloud}
\end{figure}

\begin{figure*}[h!]
    \centering
    \begin{subfigure}[b]{0.185\textwidth}
        \includegraphics[width=\linewidth]{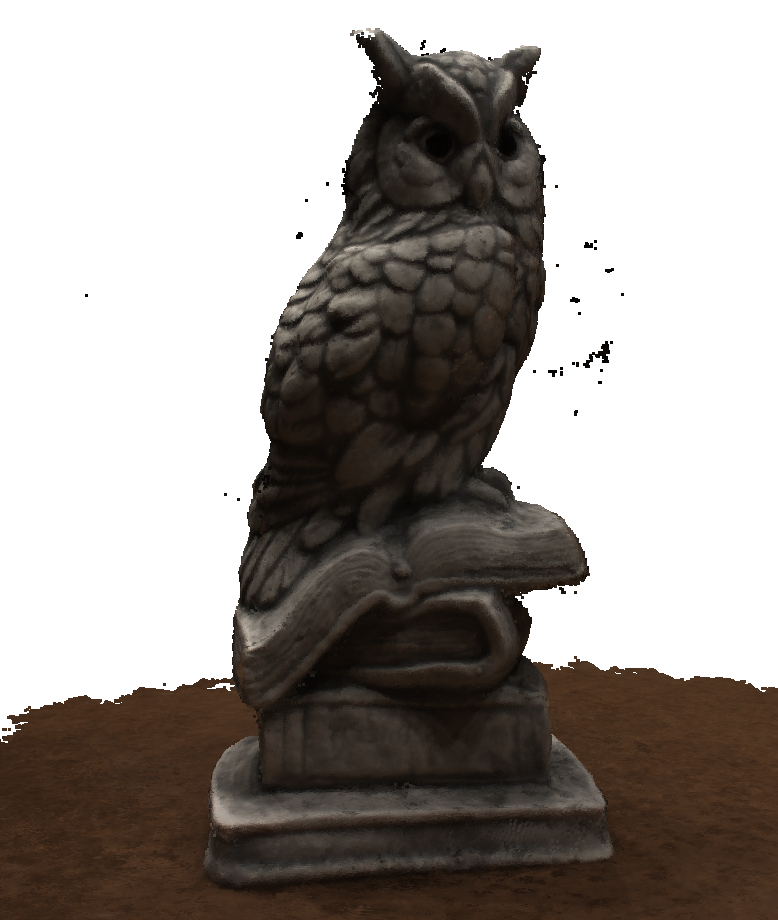}
        \caption{Input + Input}
        \label{fig:image1}
    \end{subfigure}%
    \begin{subfigure}[b]{0.185\textwidth}
        \includegraphics[width=\textwidth]{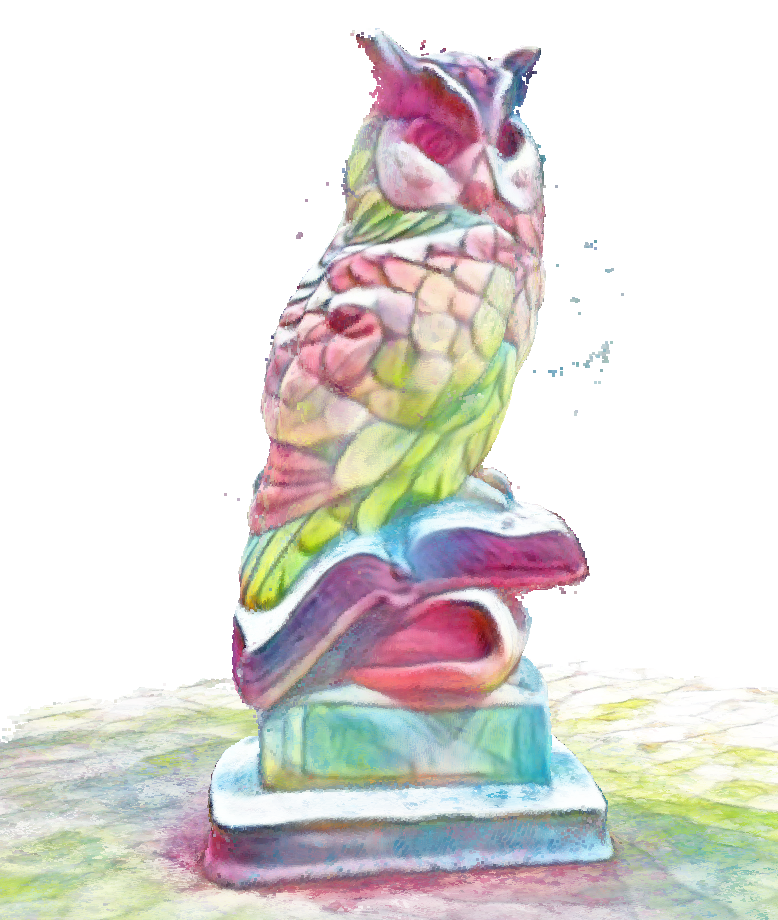}
        \caption{Input + UNet}
        \label{fig:image2}
    \end{subfigure}%
    \begin{subfigure}[b]{0.185\textwidth}
        \includegraphics[width=\textwidth]{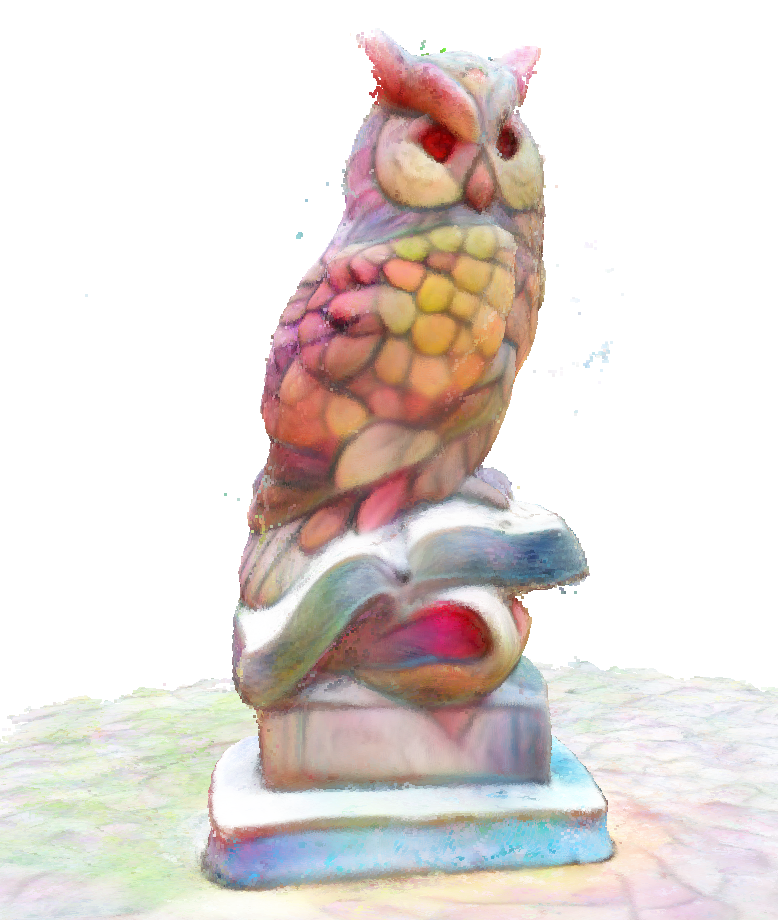}
        \caption{Input + AdaIN}
        \label{fig:image3}
    \end{subfigure}%
    \begin{subfigure}[b]{0.185\textwidth}
        \includegraphics[width=\textwidth]{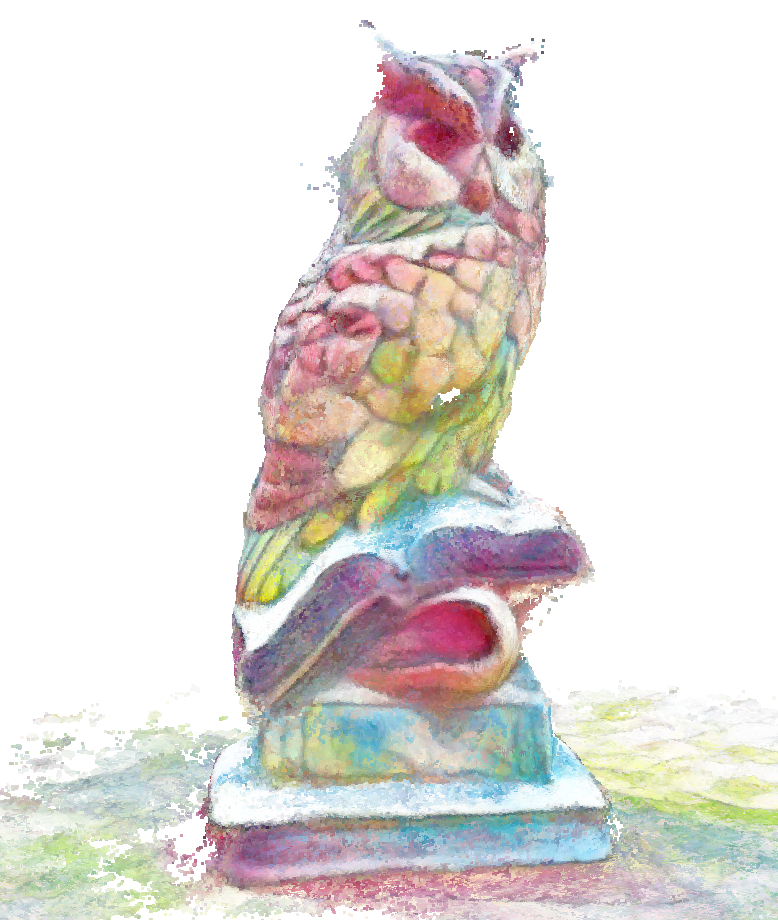}
        \caption{UNet + UNet}
        \label{fig:image4}
    \end{subfigure}%
    \begin{subfigure}[b]{0.185\textwidth}
        \includegraphics[width=\textwidth]{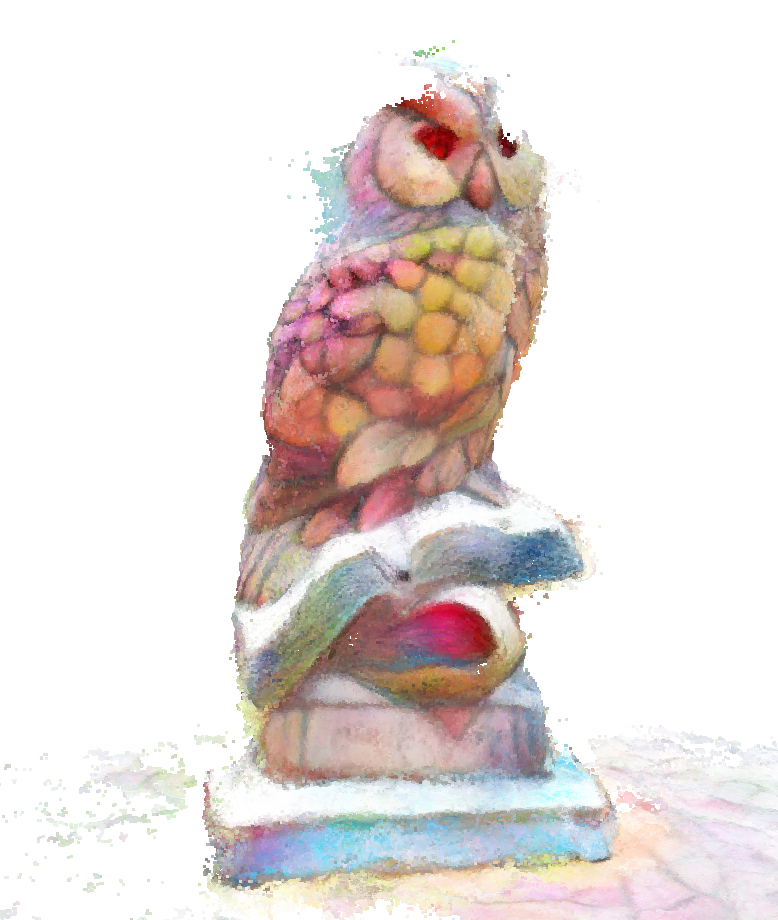}
        \caption{AdaIN + AdaIN}
        \label{fig:image5}
    \end{subfigure}
    \caption{Point cloud reconstruction results using PatchmatchNet backbone (Geometry + Coloring).   The point clouds (a), (b), and (c) are identical in terms of geometry, with (a) colored by the 64 original input images, (b) colored by the stylized images using the output of Patchmatchnet\_UNet, and (c) colored by the stylized images using the output of Patchmatchnet\_AdaIN.
    (d) and (e) are reconstructed from only the stylized images.}
    \label{fig:five_images}
\end{figure*}

Figure \ref{fig:pointcloud} shows a visual comparison of the point clouds reconstructed from the original input images and the stylized images. 
The objective of this experiment is to evaluate stylized image consistency across multiple views in this experiment. 
From the result, we can conclude that MuVieCAST  can generate highly consistent stylized multi-view images that allow us to estimate accurate depth from the styled images and yield good 3D point cloud reconstruction. 
Despite the reconstructed point cloud from stylized views exhibiting more noise in its geometry, the reconstructed point cloud remains adequate for identifying the object's geometry. 

To showcase the point cloud reconstruction using the PatchmatchNet backbone, we conducted several experiments, shown in Figure \ref{fig:five_images}. As with our previous experiment, we aim to demonstrate the consistency of the stylization across multiple views with the PatchmatchNet backbone. 
We observe that the point clouds (d) and (e) are reasonably reconstructed, where the depth maps for the two point clouds were estimated from stylized images by PatchmatchNet\_UNet and PatchmatchNet\_AdaIN, respectively. We can observe that both point clouds are sufficient for identifying the object. During our experiments, we also observed that in most cases, the UNet backbone performed better than the AdaIN backbone in terms of geometric consistency of multi-view style transfer.

\begin{figure*}[h!]
    \centering
    \begin{subfigure}[b]{0.185\textwidth}
        \includegraphics[width=\linewidth]{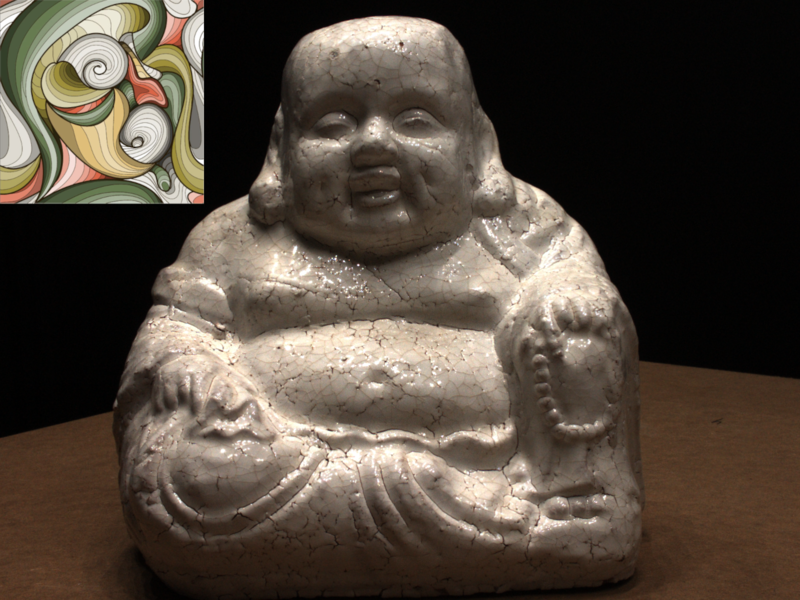}
        \includegraphics[width=\linewidth]{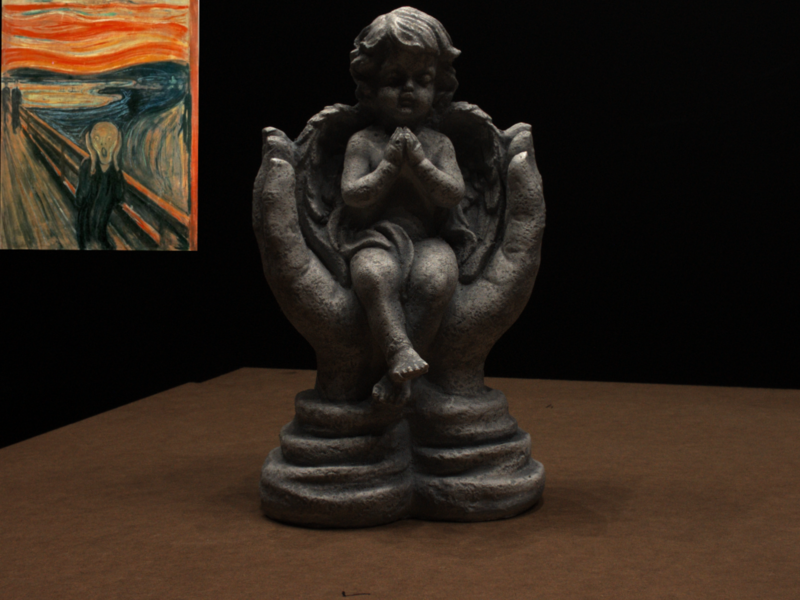}
        \includegraphics[width=\linewidth]{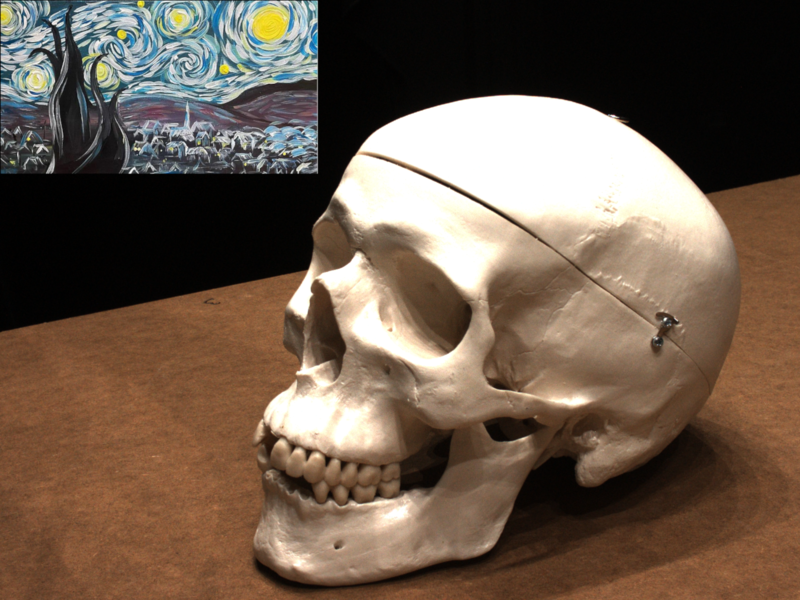}
        \includegraphics[width=\linewidth]{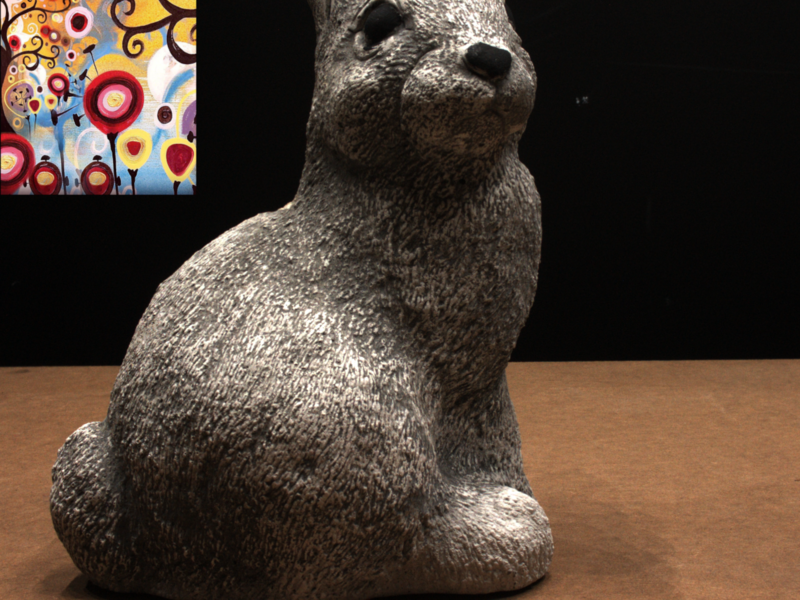}
        \caption{\\Input image}
        \label{fig:image1}
    \end{subfigure}%
    \begin{subfigure}[b]{0.185\textwidth}
        \includegraphics[width=\textwidth]{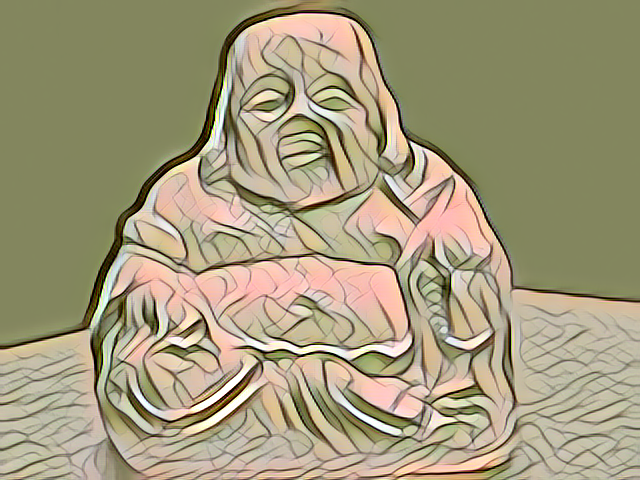}
        \includegraphics[width=\linewidth]{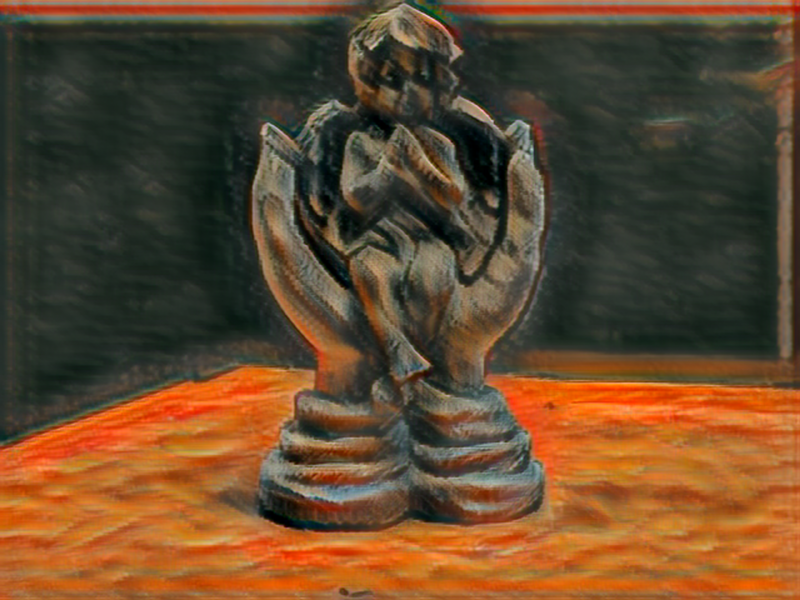}
        \includegraphics[width=\linewidth]{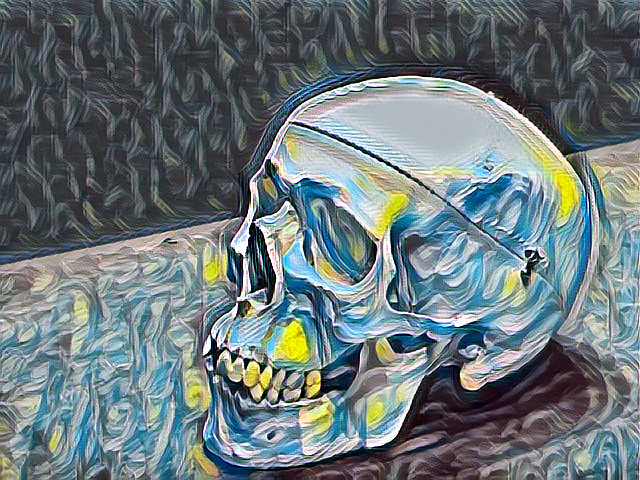}
        \includegraphics[width=\linewidth]{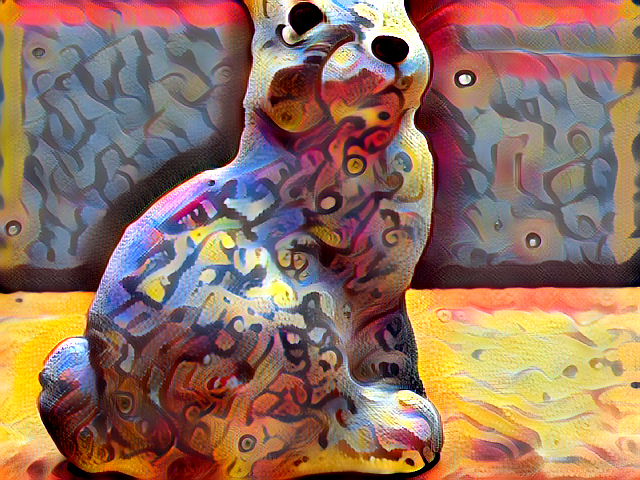}
        \caption{\\Stylized image}
        \label{fig:image2}
    \end{subfigure}%
    \begin{subfigure}[b]{0.185\textwidth}
        \includegraphics[width=\textwidth]{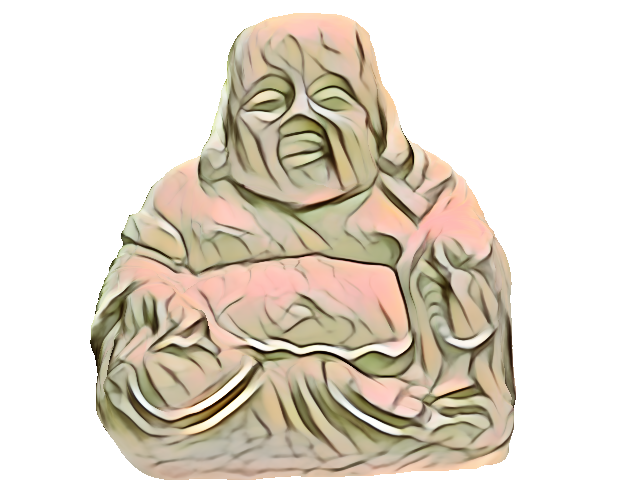}
        \includegraphics[width=\linewidth]{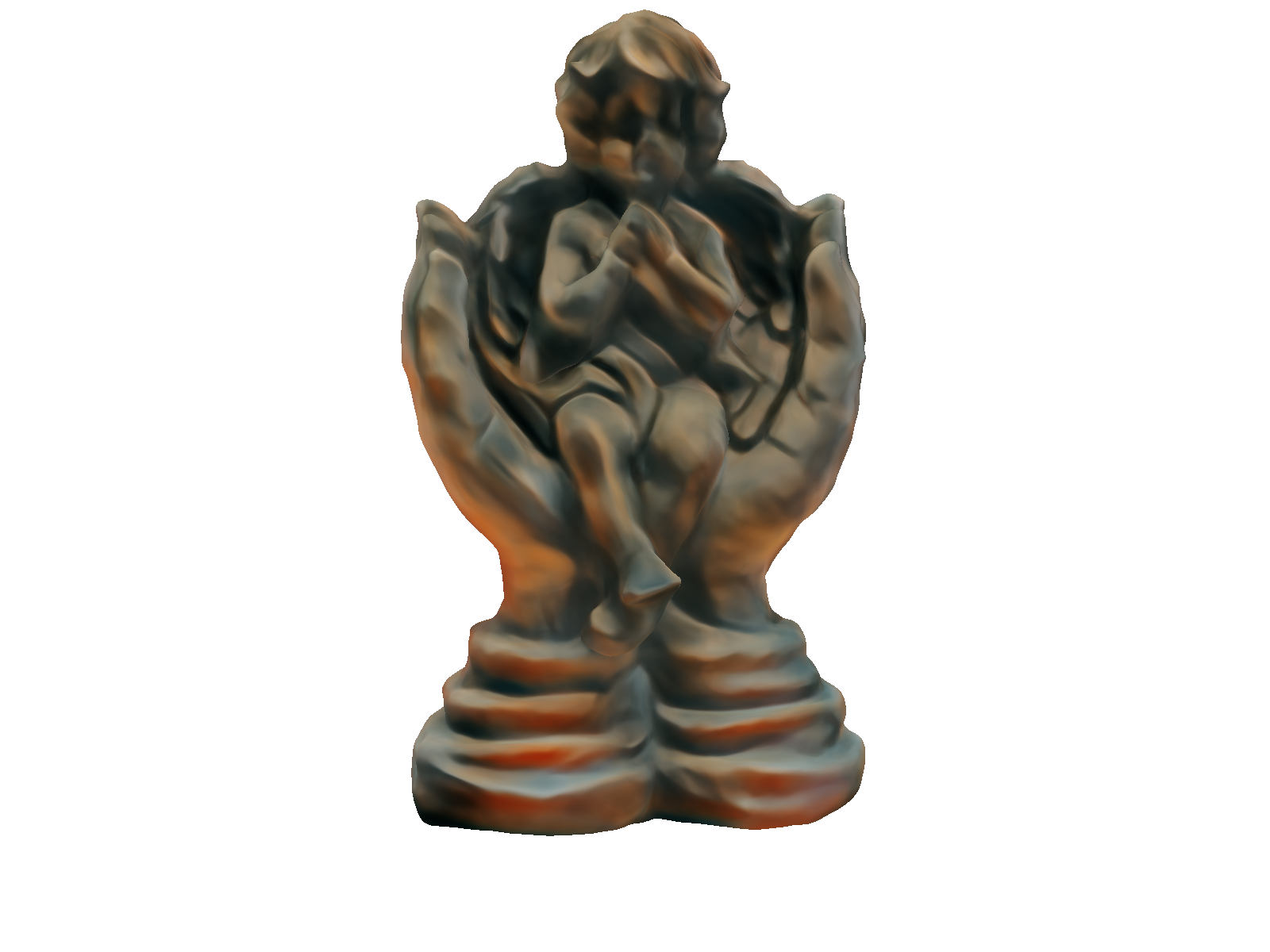}
        \includegraphics[width=\linewidth]{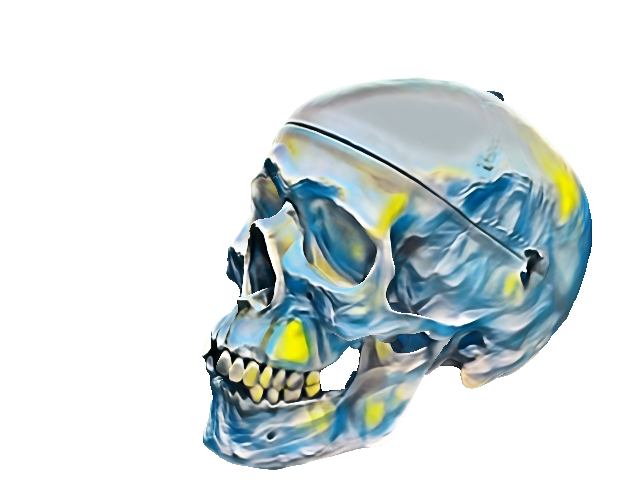}
        \includegraphics[width=\linewidth]{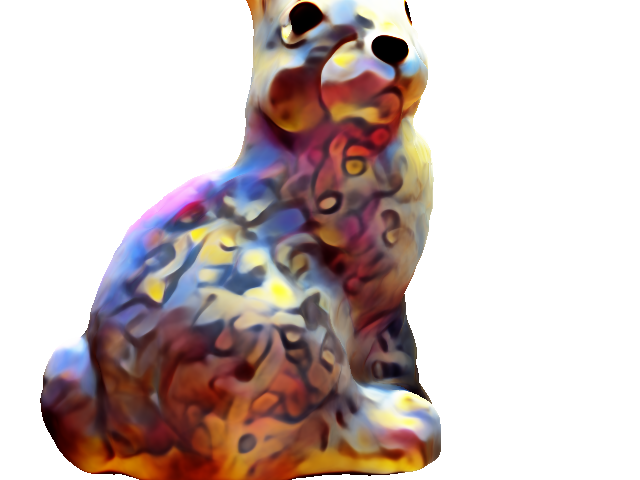}
        \caption{Stylized \\
                    surface rendering  }
        \label{fig:image3}
    \end{subfigure}%
    \begin{subfigure}[b]{0.185\textwidth}
        \includegraphics[width=\textwidth]{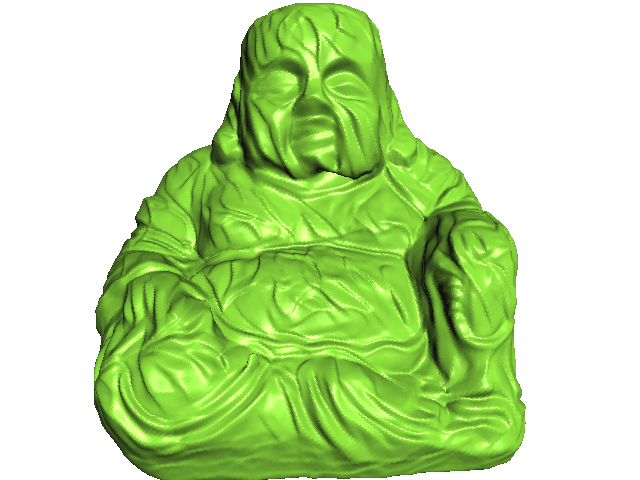}
        \includegraphics[width=\linewidth]{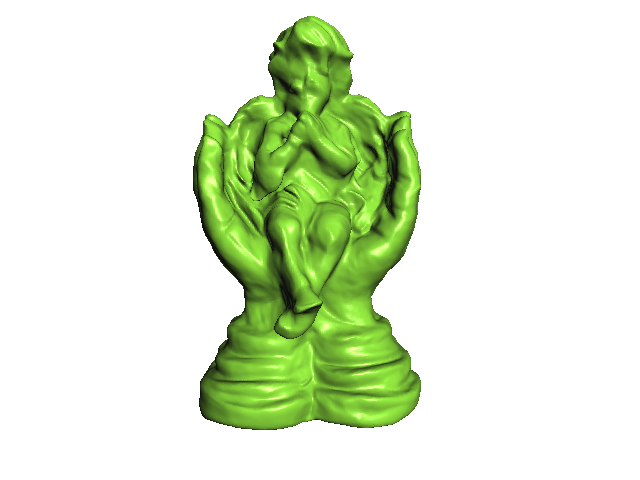}
        \includegraphics[width=\linewidth]{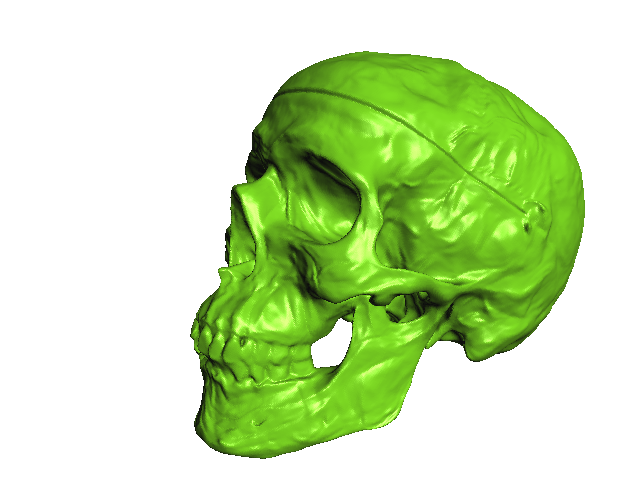}
        \includegraphics[width=\linewidth]{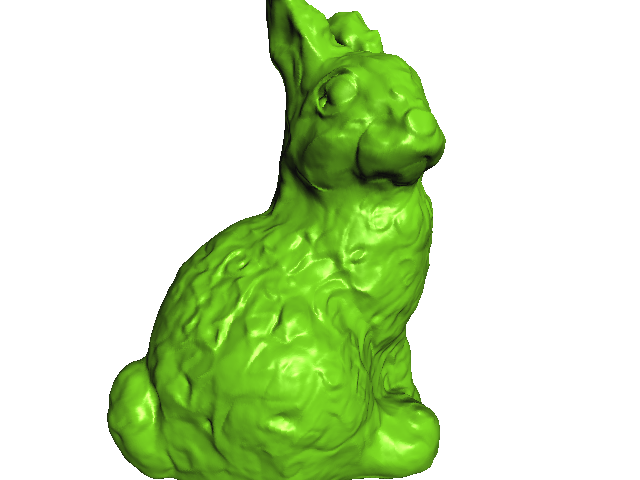}
        \caption{Reconstruction\\ from stylized images}
        \label{fig:image4}
    \end{subfigure}%
    \begin{subfigure}[b]{0.185\textwidth}
        \includegraphics[width=\textwidth]{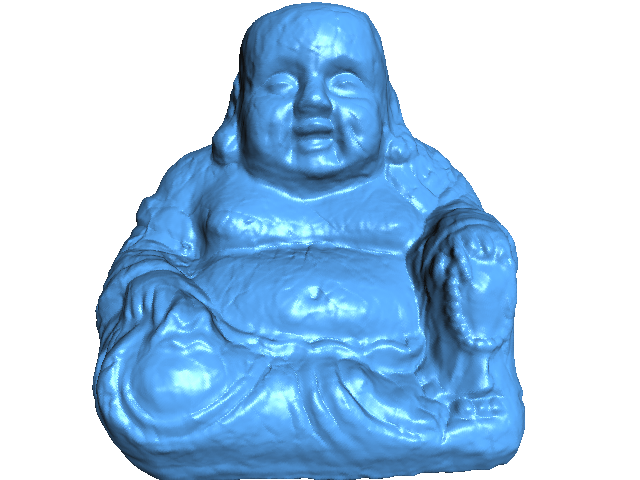}
        \includegraphics[width=\linewidth]{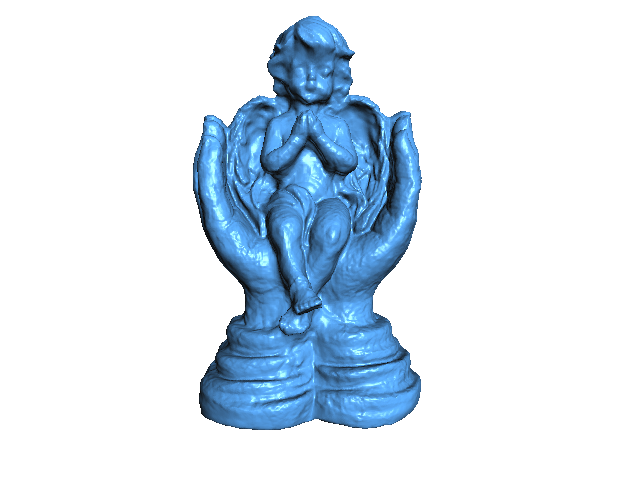}
        \includegraphics[width=\linewidth]{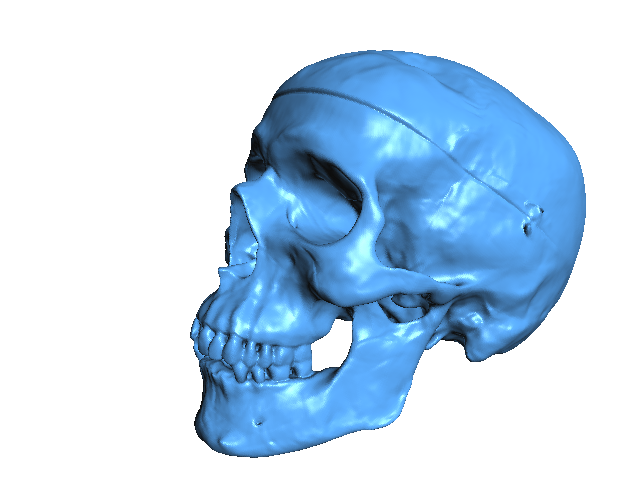}
        \includegraphics[width=\linewidth]{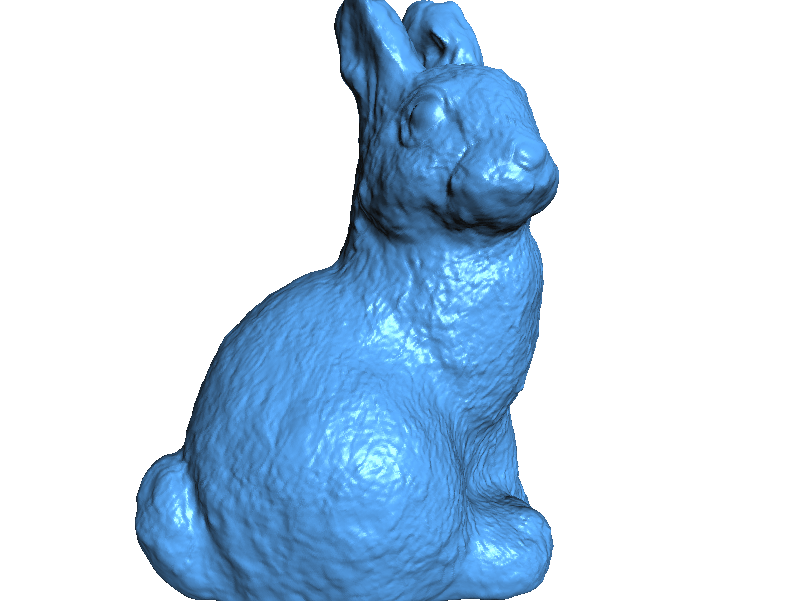}
        \caption{Reconstruction\\ from input images}
        \label{fig:image5}
    \end{subfigure}
    \caption{Neural mesh reconstruction results. First row: CasMVSNet\_UNet. Second row: CasMVSNet\_AdaIN. Third row: PatchmatchNet\_UNet. Fourth row: PatchmatchNet\_AdaIN. 
    The columns in the figure are as follows: 
    (a) Input image.  
    (b) Stylized image,
    (c) The stylized mesh surface rendering learned by IDR~\cite{yariv2020multiview}.
    (d) The mesh reconstructed from the stylized images. 
    (e) The mesh reconstruction from the original input images.}
    \label{fig:mesh_results}
\end{figure*}
\subsection{Mesh reconstruction}

In this experiment, we demonstrate the efficacy and robustness of our multi-view style transfer method for neural rendering-based mesh reconstruction~\cite{DVR,yariv2020multiview,yariv2021volume,wang2021neus}, which has gained popularity in recent years. Figure~\ref{fig:mesh_results} shows mesh reconstruction results using IDR~\cite{yariv2020multiview}. 
Our stylization method preserves consistency for mesh representation-based geometry estimation, despite some loss of finer details in the reconstructed meshes. 
These results further confirm the reliability and consistency of our robust multi-view style transfer approach, for mesh reconstruction.


 \begin{figure*}[h!]
	\centering	\includegraphics[width=0.98\textwidth]{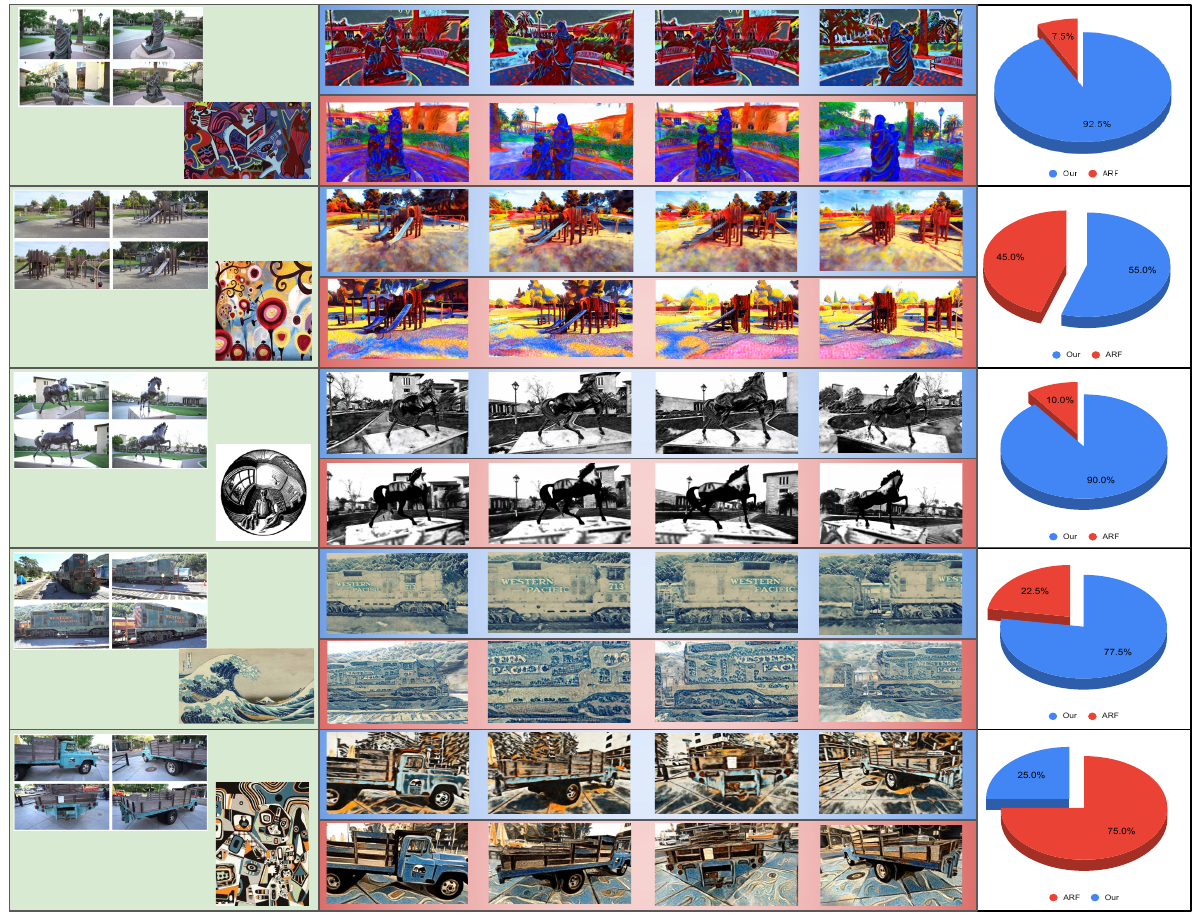}
	\caption{The left column shows the scene samples and style images shared with user study participants. In the middle column, frames from our results are presented on the top rows (on a blue stripe), while frames from the ARF method are displayed on the bottom rows (on a red stripe). The videos were provided to participants in a randomized order. The pie charts indicate the preferences of the 40 participants.}
	\label{fig:nvs}
\end{figure*}

\subsection{Novel view synthesis}

MuVieCAST offers a significant advantage over other methods~\cite{zhang2022arf,Huang22StylizedNeRF,hollein2021stylemesh} as it does not require ground truth 3D data or precomputed 3D implicit fields.  To evaluate the effectiveness of our approach compared to the recent state-of-the-art method ARF~\cite{zhang2022arf}, we conducted an anonymous user survey with 40 participants.
During the survey, participants were provided with sample scene images and style images. The evaluation was performed on five scenes from the Tanks and Temples dataset~\cite{Knapitsch2017}, each paired with a distinct style image. For each scene, participants were presented with two videos in random order. One video was generated using the ARF method, utilizing ground truth pose information and recommended parameters.

To ensure a fair assessment, we adhered closely to the default trajectory employed by ARF. Additionally, we previously demonstrated the geometric consistency of our stylization through mesh and point cloud reconstruction experiments. In these comparison experiments, we extended our evaluation beyond computing radiance fields~\cite{mildenhall2020nerf,yu2021plenoctrees,nerfstudio,mueller2022instant} to include the direct estimation of camera calibrations from stylized images using the SFM algorithm~\cite{schoenberger2016sfm}.

The results of the survey indicated that in 68\% of the cases, participants preferred our results over the ones generated by ARF. To present the survey results for each scene, Figure \ref{fig:nvs} shows the comparison outcomes.
These findings highlight the efficacy of our network architecture and its potential for practical applications in style transfer and scene generation tasks. 
\section{Conclusion}
\label{sec:conc}
We have presented MuVieCAST, a multi-view style transfer architecture, that offers a fast, versatile, and robust solution for various downstream applications, including stereo matching-based point cloud reconstruction, neural mesh reconstruction, and novel-view synthesis.
In contrast to other 3D style transfer methods, our proposed method does not require precomputed or groundtruth 3D scene representations (point cloud, mesh, radiance fields, signed distance functions, occupancy fields) and generates consistent stylized views directly from calibrated input views. 
Our experiments have demonstrated the effectiveness of different backbones in our multi-view consistent style transfer network architecture, validating our architectural ideas.
The resulting stylized images are consistent for robust geometry estimation, revealing finer geometric and texture details of the underlying 3D scene representation, thereby enhancing downstream applications. 
Moreover, our proposed network architecture trains fast, making it an appealing option for researchers and practitioners interested in related application domains.

{
    \small
    \bibliographystyle{ieeenat_fullname}
    \bibliography{main}
}
\clearpage
\setcounter{page}{1}
\maketitlesupplementary

\renewcommand\thesection{\Alph{section}}
\renewcommand\thesubsection{\thesection.\Alph{subsection}}
\setcounter{section}{0}

\section{Configuration details}
In this section, we share details of the network architecture configurations referenced in Tables~\ref{tab:backbones} and~\ref{tab:configurations} of the paper.
CasMVSNet~\cite{gu_2020_cascademvsnet} has 927K parameters, while PatchMatchNet~\cite{wang2020patchmatchnet} has 222K parameters. We divided CasMVSNet cost volume into 8 groups to calculate group-wise correlation~\cite{xu2020learning_inverse,guo2019group}.
In CasMVSNet\_UNet and PatchmatchNet\_UNet, we computed content loss at layer \textit{relu3\_3} and used the Gram matrix-based style loss ~\cite{gatys2016image} at layers \textit{relu1\_2}, \textit{relu2\_2}, \textit{relu3\_3}, and \textit{relu4\_3} of the VGG16 loss network~\cite{simonyan2014very}. To achieve this, we trimmed the VGG16 until the \textit{relu4\_3} layer, resulting in 7.6M parameters. Our UNet~\cite{ronneberger2015u} has 1.7M parameters. For CasMVSNet\_AdaIN and PatchmatchNet\_AdaIN, we computed content loss at layer \textit{relu4\_1} and style loss at layers \textit{relu1\_1}, \textit{relu2\_1}, \textit{relu3\_1}, and \textit{relu4\_1} of the VGG19 loss network. We trimmed the VGG19~\cite{simonyan2014very} until the \textit{relu4\_1} layer, which resulted in 3.5M parameters. Our trainable VGG19-based AdaIN decoder~\cite{Huang_2017_ICCV} also has 3.5M parameters.

The training time for DTU scan65 with 49 images, a resolution of $640 \times 480$, a neighboring view window size of 3, and a batch size of 1 per GPU on dual RTX 2080 Ti was measured. For various network architectures, the training times for 10 epochs are as follows:

\begin{tabular}{ |c|c| }
\hline
\textbf{Network Architecture} & \textbf{Training Time} (seconds) \\
\hline
CasMVSNet\_UNet & 174.44 \\
CasMVSNet\_AdaIN & 174.52 \\
PatchmatchNet\_UNet & 153.03 \\
PatchmatchNet\_AdaIN & 155.00 \\
\hline
\end{tabular}

\section{UNet style translation}
\begin{figure}[!h]
  \centering
  \includegraphics[width=0.32\columnwidth]{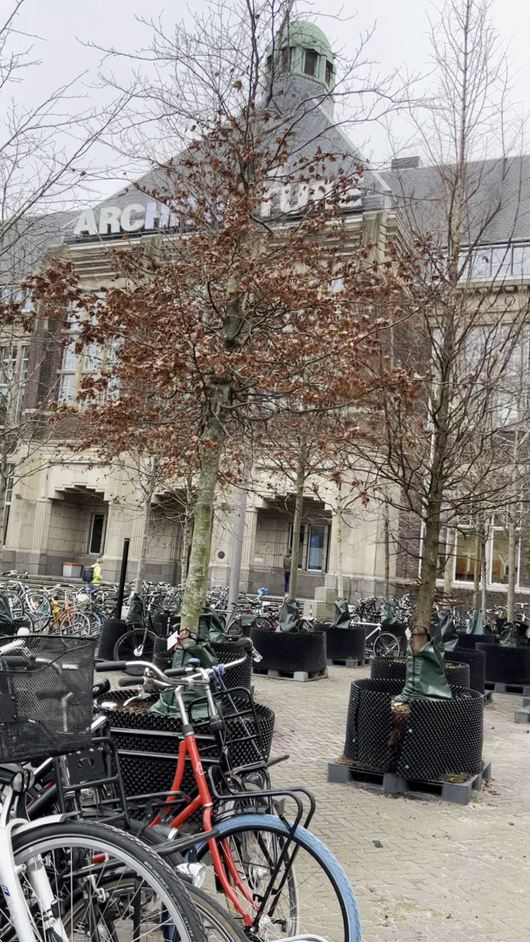}
  \includegraphics[width=0.32\columnwidth]{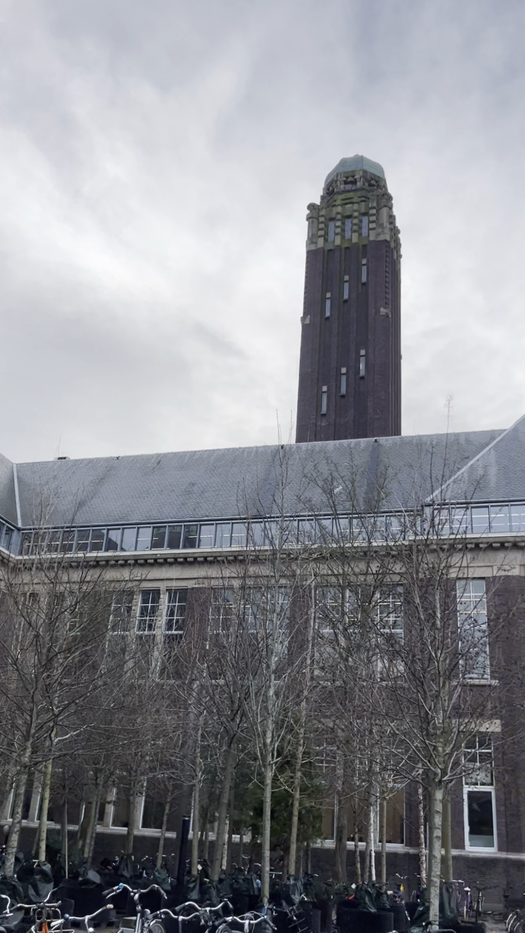}
  \includegraphics[width=0.32\columnwidth]{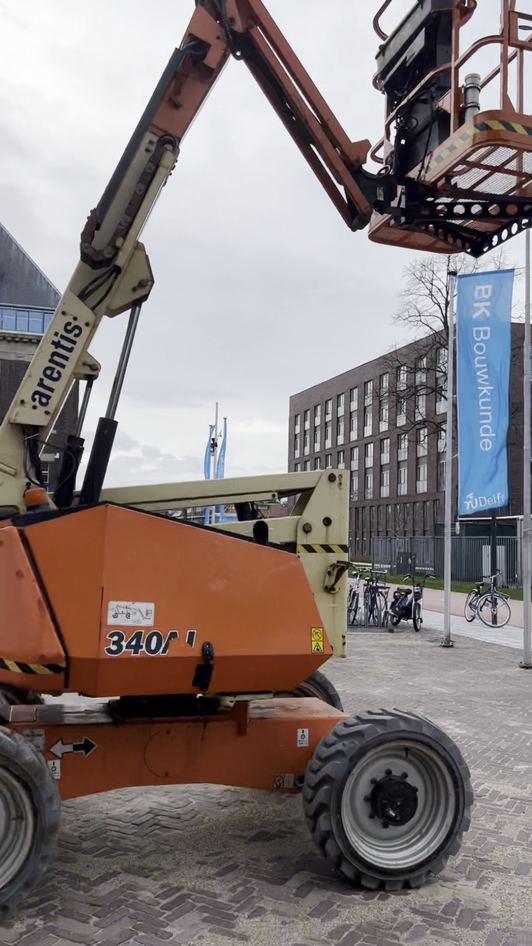}\\
  \includegraphics[width=0.32\columnwidth]{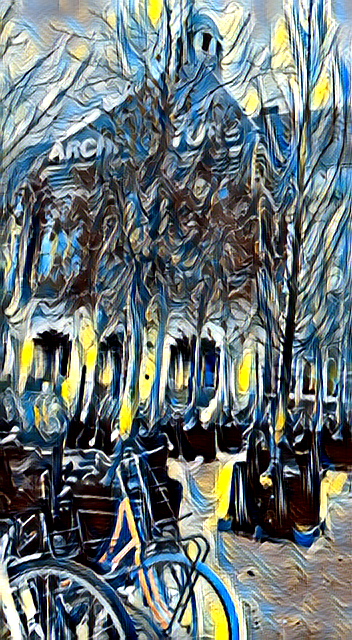}
  \includegraphics[width=0.32\columnwidth]{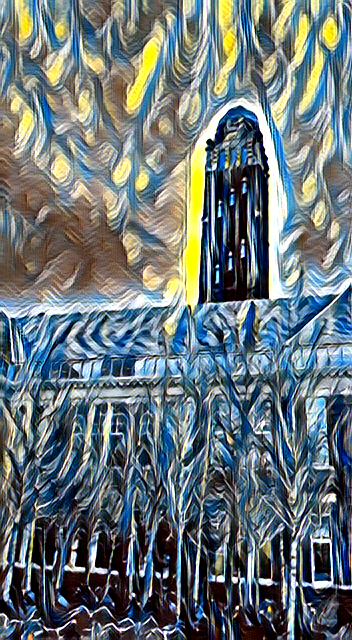}
  \includegraphics[width=0.32\columnwidth]{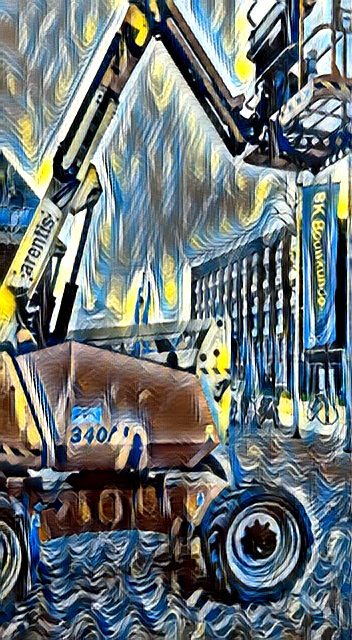}\\
  \includegraphics[width=0.32\columnwidth]{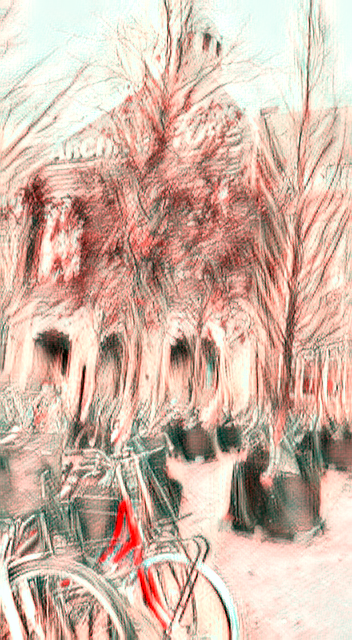}
  \includegraphics[width=0.32\columnwidth]{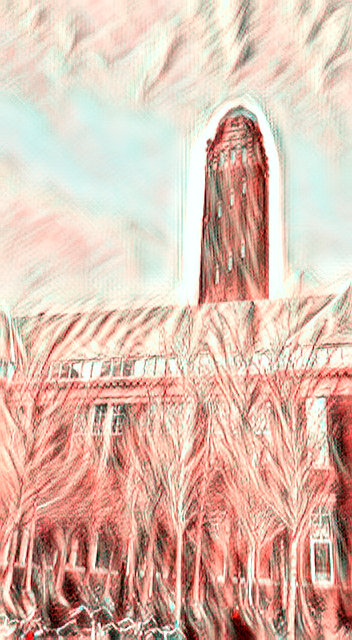}
  \includegraphics[width=0.32\columnwidth]{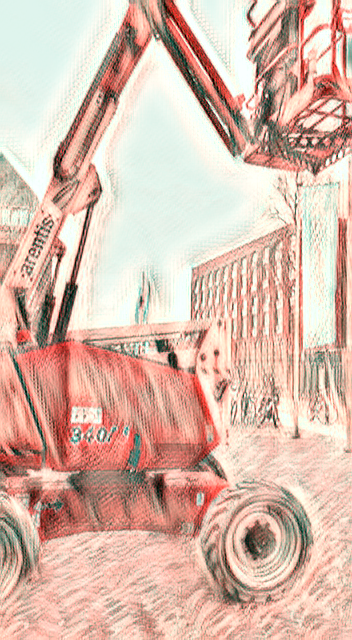}\\
  \includegraphics[width=0.99\columnwidth]{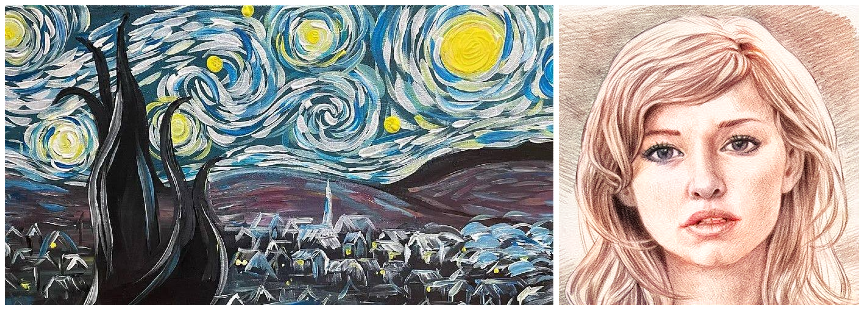}\\
  \caption{Results for style translation. First row: original input samples. Second row: results with pretrained style. Third row: finetuning with a new style image. Bottom row: the two style images used in the second and third rows, respectively.}
  \label{fig:style_translation}
\end{figure}

Although UNet-based style transfer methods~\cite{Johnson2016Perceptual} require style-specific pretraining, we observed that UNet's pretrained models can quickly adapt to new styles and images. 
By using weights learned from MS COCO object detection dataset~\cite{lin2014microsoft}, we can swiftly apply UNet to new style images for given inputs. 
Pretraining captures a general understanding of style features and textures that can be reused, significantly reducing the time and resources needed for training new models for each style.
In this experiment, we used pretrained model of UNet trained with the ``Starry Night". We pretrained the model with a new hand-drawn style image~\cite{Huang_2017_ICCV,zhang2017multistyle} for 30 mins with a dual RTX2080 Ti. Figure~\ref{fig:style_translation} depicts results for style translation.

\section{Color adjustment}

\begin{figure}[htbp]
  \centering
  \includegraphics[width=0.45\columnwidth]{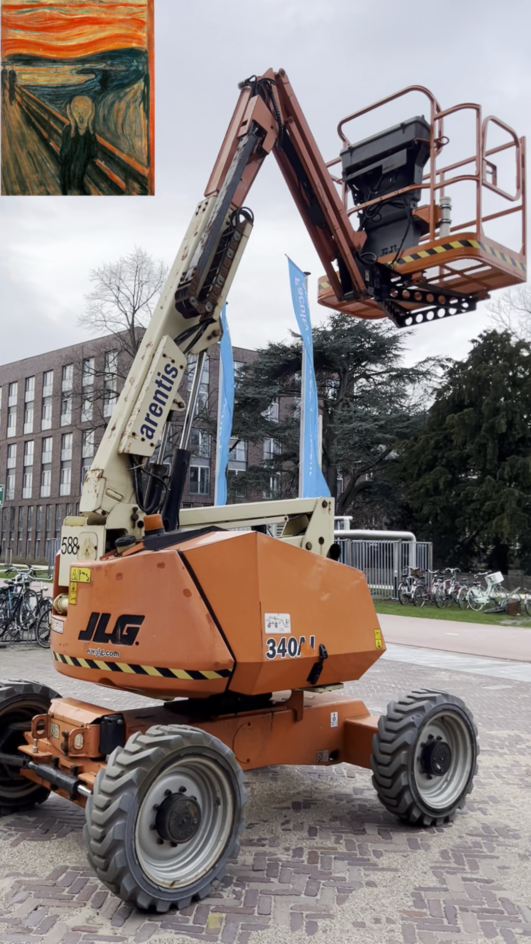}
  \includegraphics[width=0.45\columnwidth]{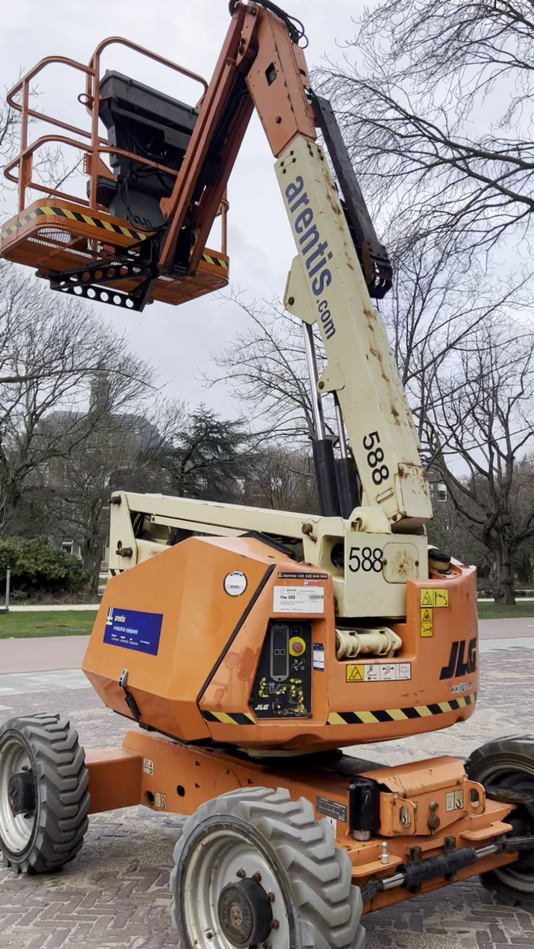}\\
  \includegraphics[width=0.45\columnwidth]{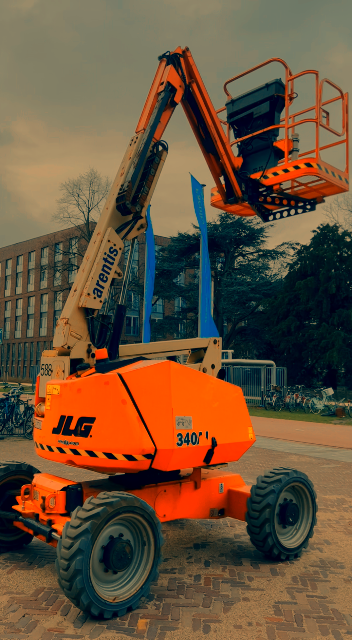}
  \includegraphics[width=0.45\columnwidth]{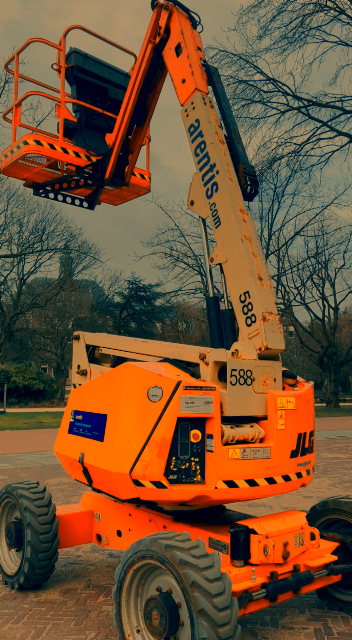}
  \caption{Color adjustment as preprocessing. 
Top: original images. Bottom: after color adjustment}
  \label{fig:pre_color}
\end{figure}
\begin{figure}[htbp]
  \centering
  \includegraphics[width=0.445\columnwidth]{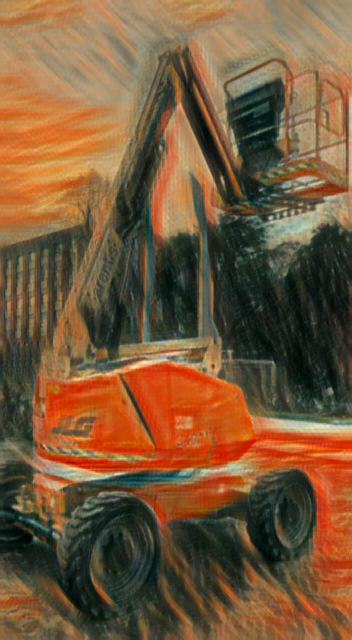}
  \includegraphics[width=0.445\columnwidth]{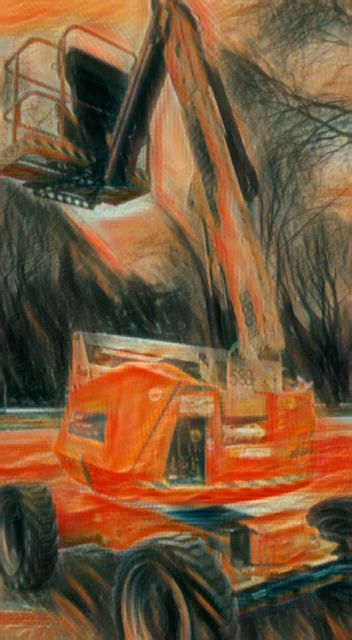}\\
  \includegraphics[width=0.445\columnwidth]{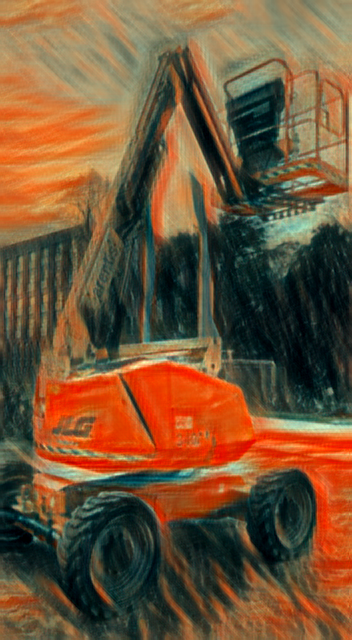}
  \includegraphics[width=0.445\columnwidth]{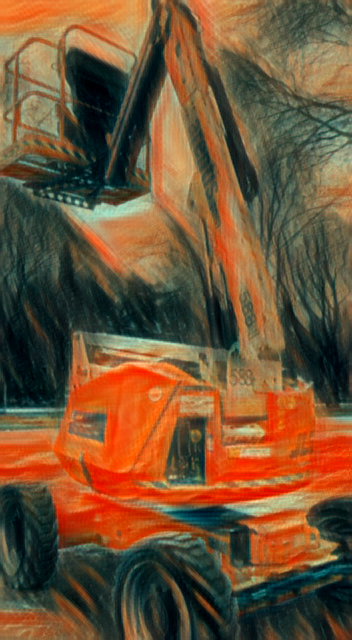}
  \caption{Color adjustment as postprocessing. 
Top: styled images. Bottom: after color adjustment}
  \label{fig:post_color}
\end{figure}
Our framework includes a color adjustment feature that allows users to modify the color of images so that their color distribution is more similar to a specified style image. Similar approaches have been previously explored in the literature~\cite{WCT-NIPS-2017,zhang2022arf}. Our framework supports both pre-processing and post-processing color adjustment. In the paper, we have only used color adjustment for novel view synthesis experiments, as both a pre-processing step and a post-processing step.

Our approach is inspired by the work of WCT~\cite{WCT-NIPS-2017}, where we aim to match the RGB color means and covariances of the input images with those of a style image. This color mapping can be represented as an affine transformation
\begin{equation}
\Tilde{c} = Mc+t,
\end{equation}
where $c$ represents the input color that is transformed into $\Tilde{c}$. $M$ is responsible for translating color vectors between the content and style domains, which is computed as $U_s\lambda_s^\frac{1}{2}V_s^TU_c\lambda_c^{-\frac{1}{2}}V_c^T$. $U_c$, $\lambda_c$, and $V_c$ are computed using singular value decompositions of the colors of the content images, while $U_s$, $\lambda_s$, and $V_s$ are computed using singular value decompositions of the colors of the style image. The translation vector $t$ adjusts the mean and can be computed as $\mu_s-M\mu_c$, where $\mu_c$ is the mean color of the content images and $\mu_s$ is the mean color of the style image.

As the color mapping is affine, the color adjustment module can be used as either a pre-processing step or a post-processing step for downstream applications related to geometry. 
Figure \ref{fig:pre_color} and Figure \ref{fig:post_color} demonstrate the effect of using color adjustment as preprocessing and postprocessing, respectively.

\begin{figure*}[hbt!]
    \centering
    \begin{subfigure}[t]{0.98\textwidth}
        \includegraphics[width=\linewidth]{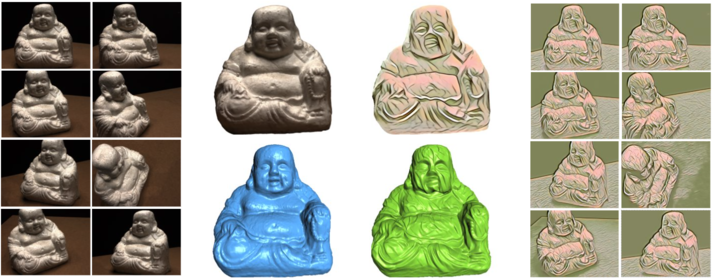}
        \label{fig:top}
    \end{subfigure}\\ 
    \begin{subfigure}[t]{0.245\textwidth}
        \includegraphics[width=\linewidth]{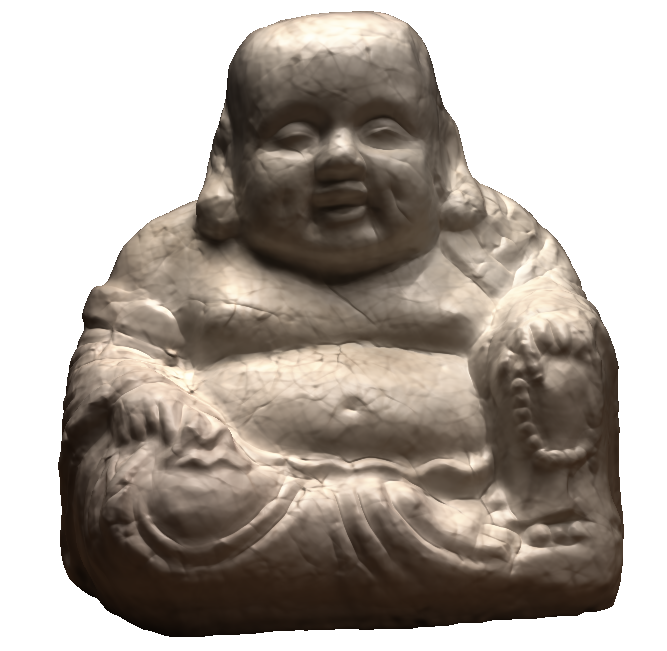}
        \caption{Input + Input}
        \label{fig:image1}
    \end{subfigure}%
    \begin{subfigure}[t]{0.245\textwidth}
        \includegraphics[width=\textwidth]{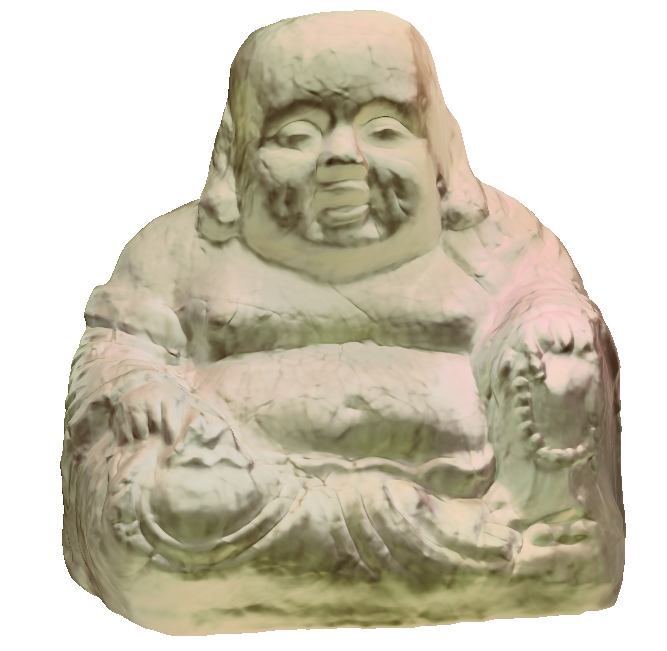}
        \caption{Input + Stylized}
        \label{fig:image2}
    \end{subfigure}%
    \begin{subfigure}[t]{0.245\textwidth}
        \includegraphics[width=\textwidth]{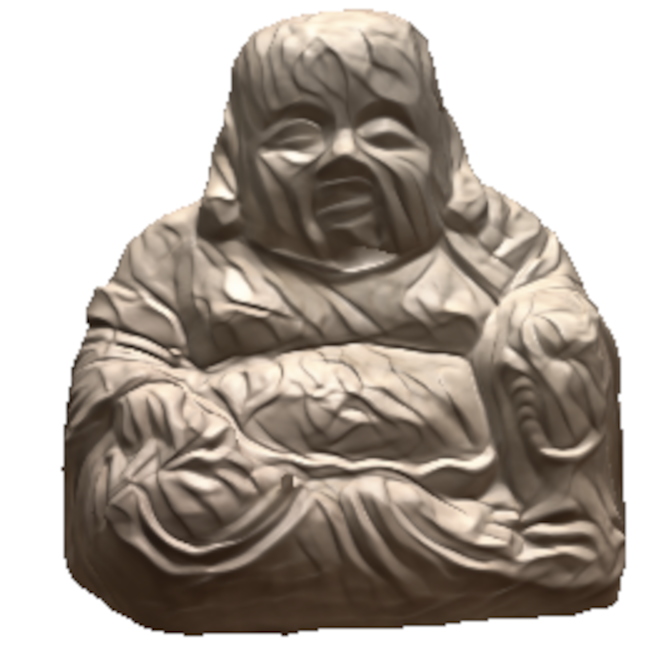}
        \caption{Stylized + Input}
        \label{fig:image3}
    \end{subfigure}%
    \begin{subfigure}[t]{0.245\textwidth}
        \includegraphics[width=\textwidth]{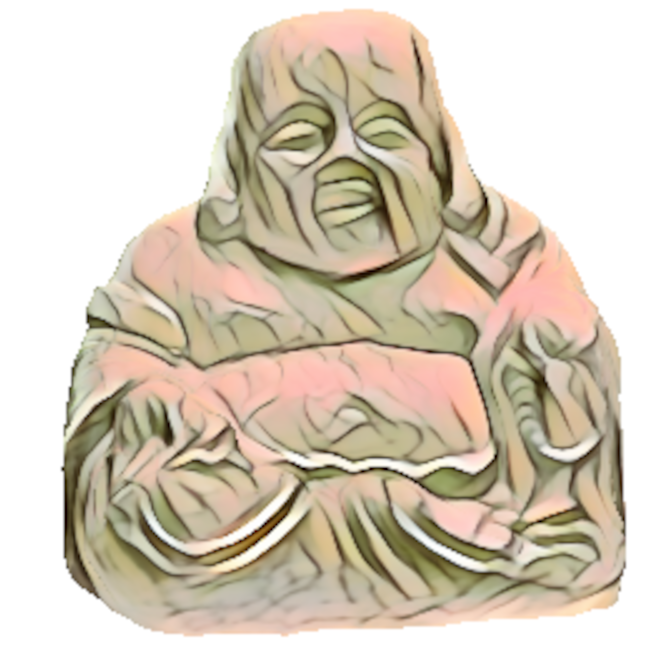}
        \caption{Stylized + Stylized}
        \label{fig:image4}
    \end{subfigure}%
    \caption{Neural mesh reconstruction.
    Top row: Mesh editing effect. The reconstructed mesh demonstrates stripe-like geometric features.\\
    Bottom row: Mesh surface coloring (Geometry + Coloring).  Meshes (a) and (b) have the same geometric properties but differ in their coloring. Specifically, (a) and (c) are colored using 64 original input images, while (b) and (d) are colored using the stylized images. The geometry of (a) and (b) is derived from the original inputs, while the geometry of (c) and (d) is learned from the stylized images.}
    \label{fig:mesh_edit_color}
\end{figure*}
\FloatBarrier
\section{Neural mesh reconstruction}


By being independent of the 3D representation of the input, MuVieCAST has the potential to be used and extended as a tool for editing 3D scene representations. By using consistent 2D stylized images, we can generate 3D geometric textures. The top row of Figure \ref{fig:mesh_edit_color} shows such an example, where the reconstructed mesh demonstrates stripe-like geometric features generated from the stylized images. These consistent features across multiple views are reflected in the reconstructed mesh as a geometric texture.

Our approach can also be applied to mesh coloring, similar to the experiments on point clouds. Recently, there are neural mesh rendering techniques~\cite{yariv2020multiview,yariv2021volume,wang2021neus} that disentangle geometry reconstructions from view-dependent appearance estimations. Leveraging this architectural advantage, we can utilize our method to color the meshes. We trained the geometry and rendering network of IDR~\cite{yariv2020multiview} using both the original images and the stylized images. The bottom row of Figure~\ref{fig:mesh_edit_color} demonstrates our experiment results for mesh coloring.

 \section{Novel-view synthesis with real-world data}
 To further assess the capabilities and robustness of MuVieCAST, we conducted novel-view synthesis experiments using real-world data. Similar to our previous Nerf experiments, we derived camera poses from stylized images to show the robustness of our method. 
Figure \ref{fig:nvs}, the left column shows parts of the original input and stylized images along with the style image.
Our proposed stylization approach preserves the essential structure and details of the scene while introducing a unique artistic style, as shown by these stylized images.
Moreover, the right column shows the novel view synthesis (NVS) results using the computed radiance fields and camera parameters from stylized images. 
The novel views exhibit fine details, accurate geometry, and coherent lighting with stylized rendering.

 \begin{figure*}[htbp]
 	\centering
 	\includegraphics[width=0.99\textwidth]{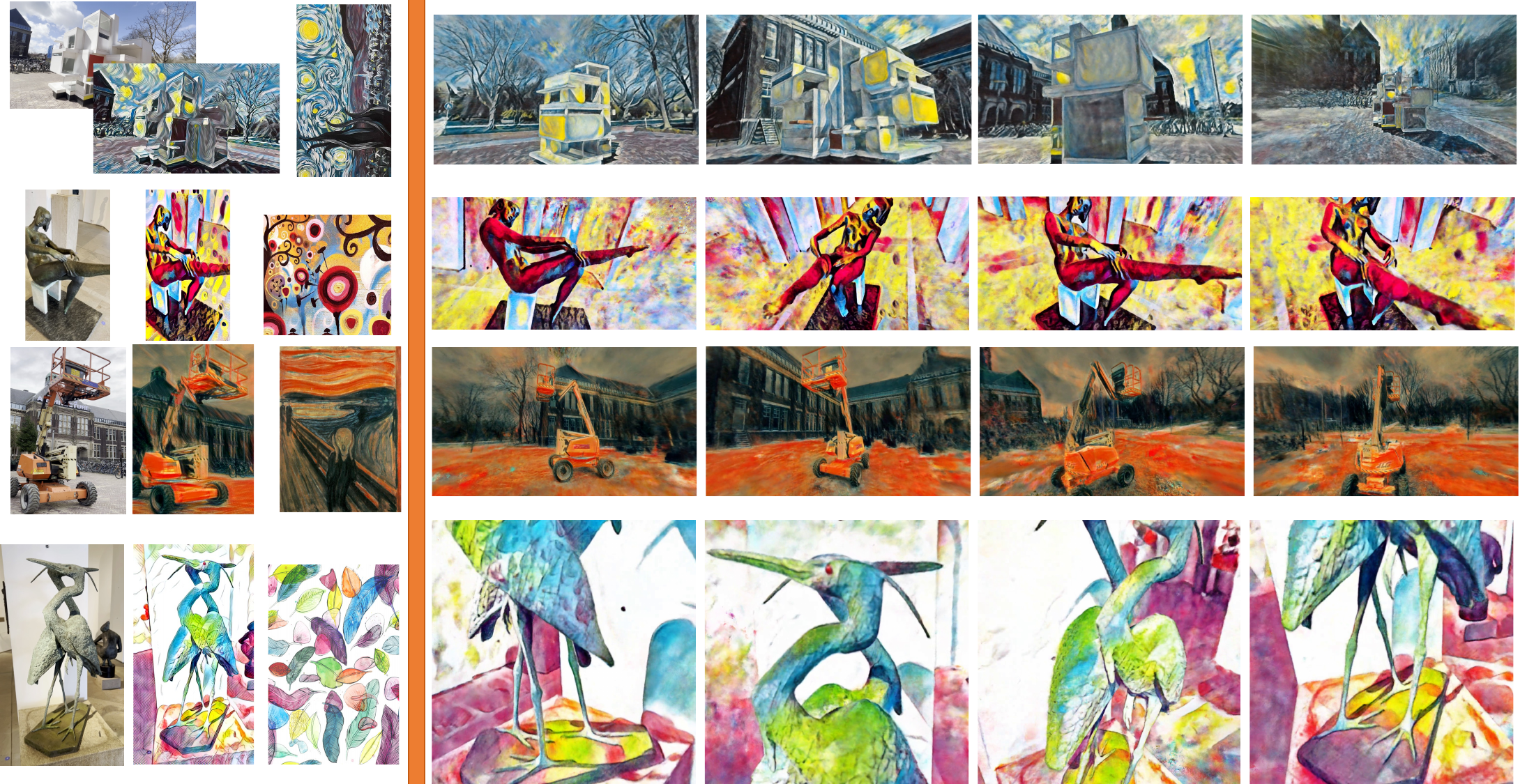}
 	\caption{Novel view synthesis using four distinct styles, applied to real-world data. MuVieCAST was trained on each scene for less than 15 minutes, leveraging a dual RTX 2080 GPU. More results are demonstrated on our project page: \href{https://muviecast.github.io/}{muviecast.github.io} }
 	\label{fig:nvs}
 \end{figure*}

\section{Loss terms} 

During the development process, we conducted over 500 automated TensorBoard experiments to define the losses.

Our empirical findings indicate that substituting SmoothL1 loss for MSE loss in the context of content loss and interchanging MSE loss with SmoothL1 loss for image geometry loss terms yields comparable outcomes. 
We extensively investigated different approaches for volume loss, including KL divergence, MSE, SmoothL1, and custom losses. Our findings demonstrated that SmoothL1 loss produced better learning curves. 
It is important to note that achieving convergence in volume loss without considering depth loss can be challenging.
Regarding image geometry loss terms, both MSE and SmoothL1 losses performed
similarly. 

In addition to the aforementioned loss terms, we also experimented with some other losses, such as nearest-neighbor-based losses~\cite{li2016combining,zhang2022arf} and total variations~\cite{Johnson2016Perceptual}, similar to previous works. However, our experiments with nearest-neighbor-based feature matching loss did not result in proper improvement, and it proved to be computationally expensive. Our experiments with total-variation (TV) loss did not generate visually and geometrically better results either. Despite this, our framework still supports these loss terms (nearest-neighbor feature matching NNFM~\cite{zhang2022arf} and total-variation~\cite{Johnson2016Perceptual}) for users who want to conduct further experiments.

In Sec.~\ref{sec:ablation}, we conduct an ablation study involving loss terms within a sparse view scenario.
\section{User survey results}
\begin{figure*}[b]
    \centering
    \begin{subfigure}{0.93\textwidth}
        \includegraphics[width=\textwidth]{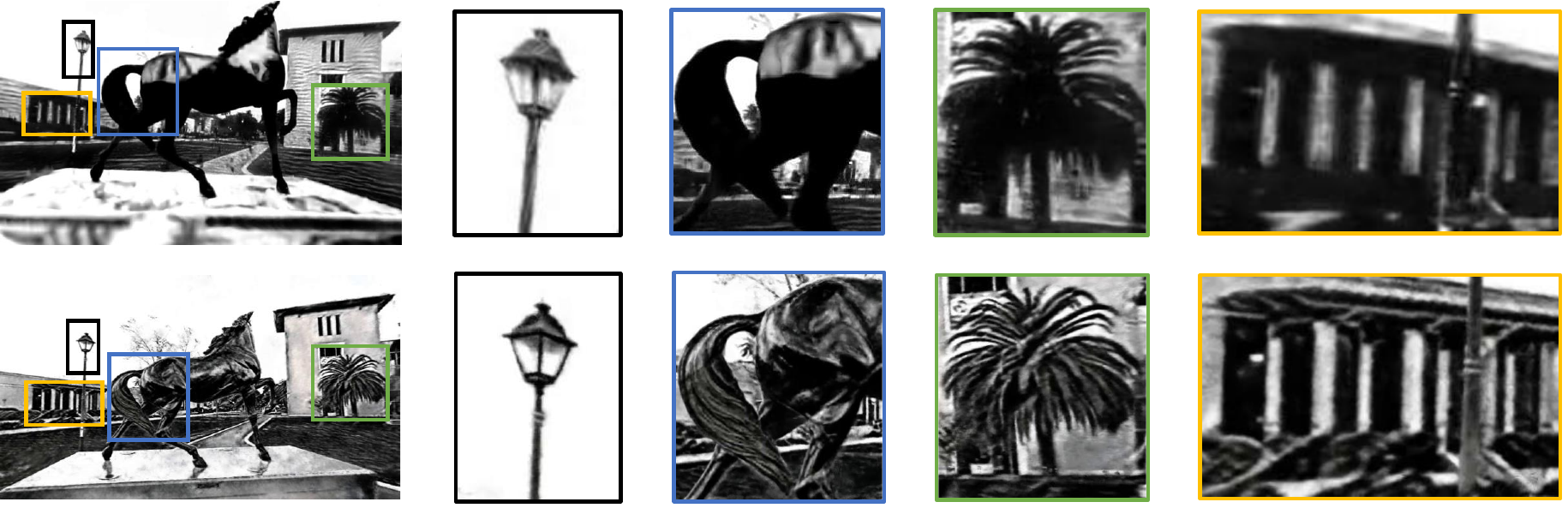}
        \caption{36 out of 40 participants chose our output (second row). }
        \label{fig:nvs3}
    \end{subfigure}\\%
    \begin{subfigure}{0.95\textwidth}
        \includegraphics[width=\textwidth]{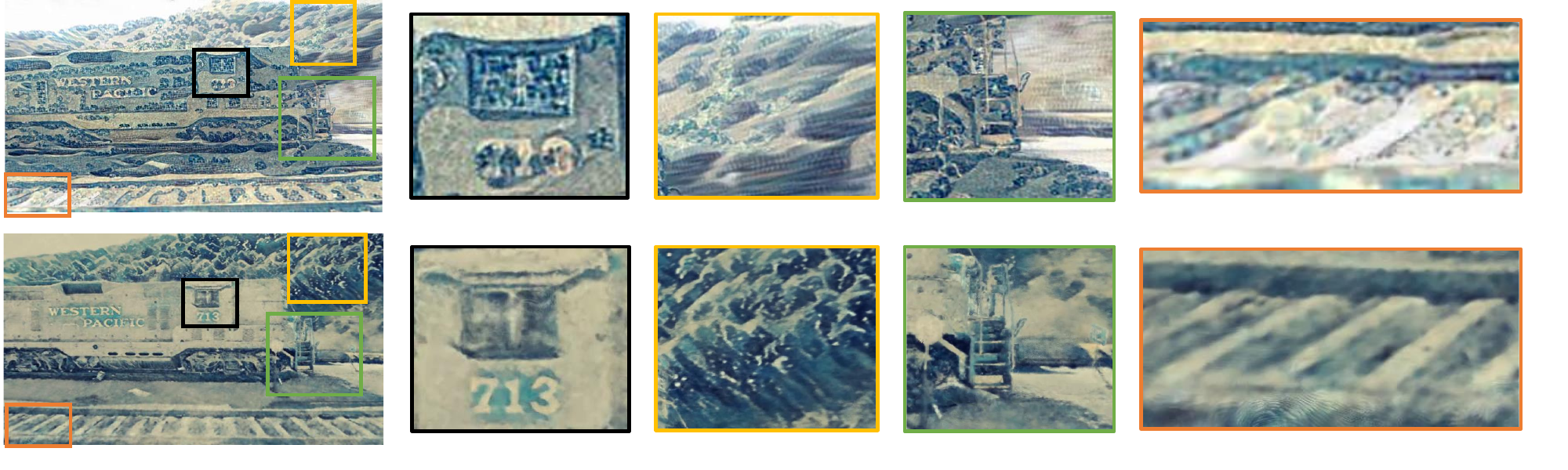}
        \caption{31 out of 40 participants chose our output (second row). }
        \label{fig:nvs4}
    \end{subfigure}\\%
\end{figure*}
We conducted a user survey involving 40 participants across 5 scenes and 5 styles, following the trajectory set by the author of ARF~\cite{zhang2022arf}. Here, we aim to delve into the reasoning behind the outcomes of our user study. We specifically showcase paired renders per scene: one from ARF and one from our model. It is important to note that our approach involves single-step optimization for multi-view consistent style transfer, and our radiance fields utilize estimated camera poses from styled images.

Here we aim to interpret the factors contributing to the outcomes of the user survey. 
Figure \ref{fig:zoomed_comparison} provides a closer examination of the comparison between our results and those of ARF. 
In Figure~\ref{fig:zoomed_comparison} (a), our results were preferred by 90\% of participants (36 out of 40) because of their richer texture compared to ARF. Figure~\ref{fig:zoomed_comparison} (b) illustrates that 77.5\% of votes (31 out of 40) leaned toward our interpretation as our style transfer maintained geometric and semantic clarity, unlike ARF, which introduced noisy artistic textures, complicating scene understanding. Moving to Figure~\ref{fig:zoomed_comparison} (c), our approach produced smoother and more visually appealing coloring, preserving semantic and geometric understanding, whereas ARF resulted in patchy high-frequency coloring. This example received 92.5\% of the votes (37 out of 40). Figure~\ref{fig:zoomed_comparison} (d) showed a more balanced preference, with a slight inclination (55\% of the votes) toward our results. Upon inspection, our method excelled in recovering distant objects, while ARF rendered sharper depictions of ground and foreground elements. Lastly, in Figure~\ref{fig:zoomed_comparison} (e), 75\% of participants (30 out of 40) opted for ARF results. Our analysis indicated that our results appeared more blurred compared to ARF, highlighting sharper radiance fields in the latter.

\begin{figure*}[tp]
    \ContinuedFloat
    \centering
    \begin{subfigure}[t]{0.99\textwidth}
        \includegraphics[width=\linewidth]{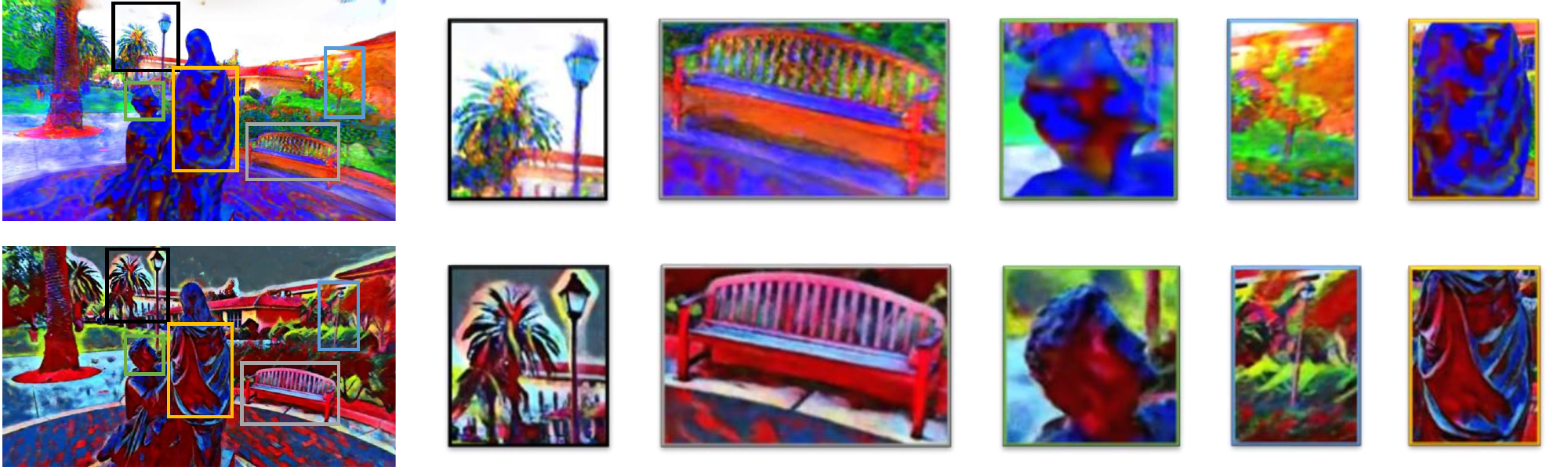}
        \caption{37 out of 40 participants chose our output (second row).}
        \label{fig:nvs1}
    \end{subfigure}\\%
    \begin{subfigure}[t]{0.99\textwidth}
        \includegraphics[width=\textwidth]{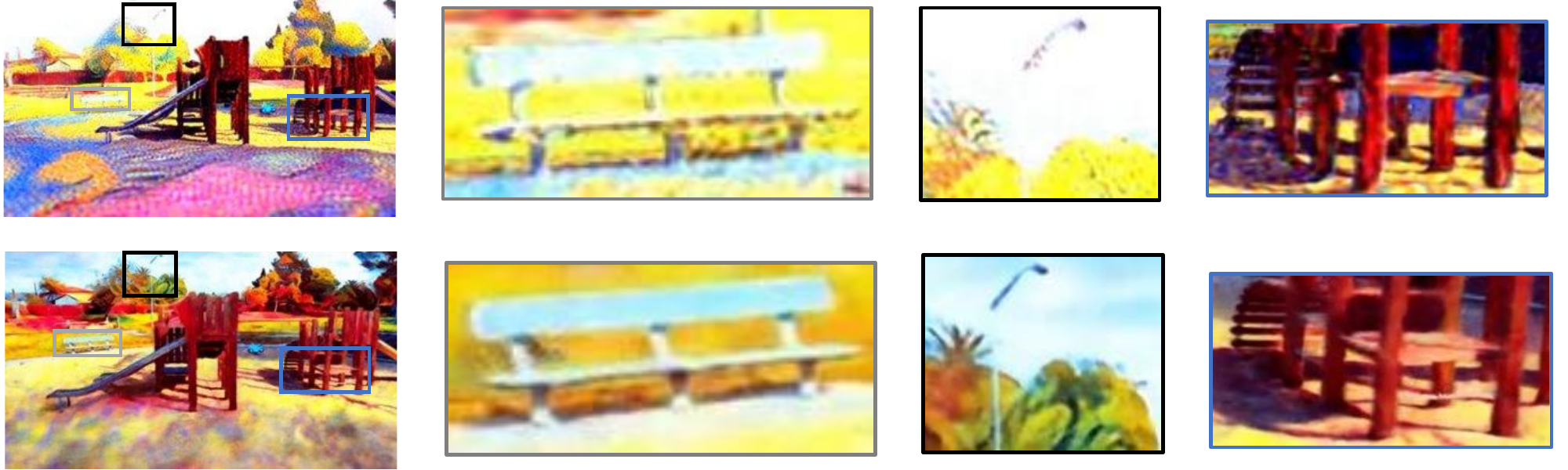}
        \caption{22 out of 40 participants chose our output (second row).}
        \label{fig:nvs2}
    \end{subfigure}\\%
    
    \begin{subfigure}[t]{0.99\textwidth}
        \includegraphics[width=\textwidth]{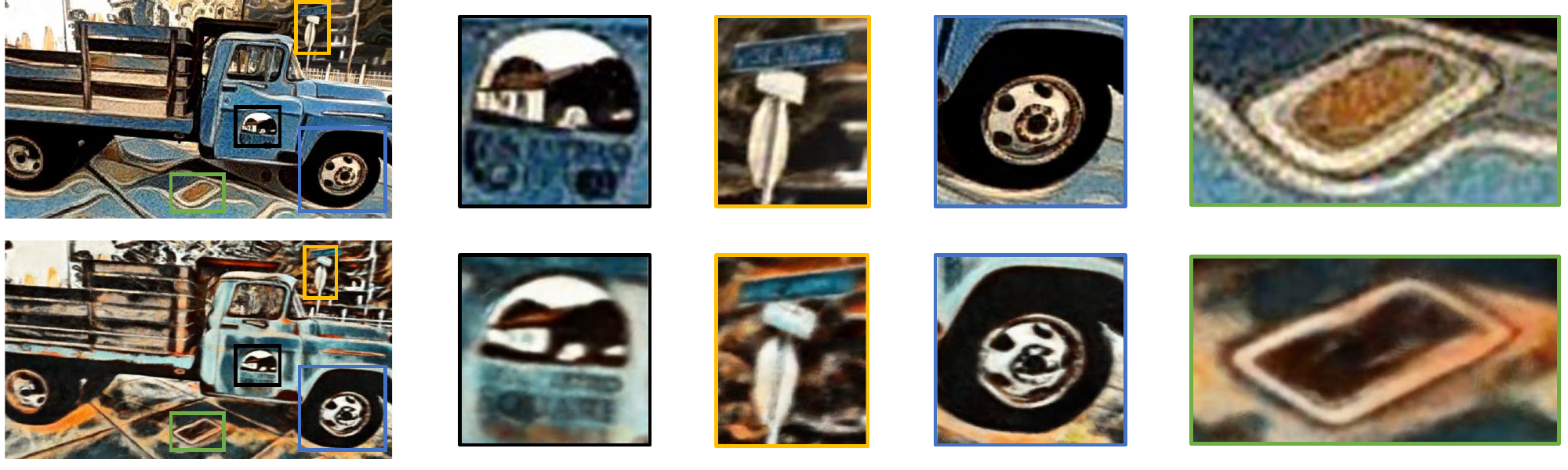}
        \caption{30 out of 40 participants chose ARF output (first row). }
        \label{fig:nvs5}
    \end{subfigure}%
    \caption{Comparison between ARF~\cite{zhang2022arf} and our results. The first row in each subfigure shows the ARF outcome, while our result is given in the second row.}
    \label{fig:zoomed_comparison}
\end{figure*}

\section{Ablation study for sparse views}
\label{sec:ablation}
\begin{figure*}[h!]
	\centering
	\includegraphics[width=0.90\textwidth]{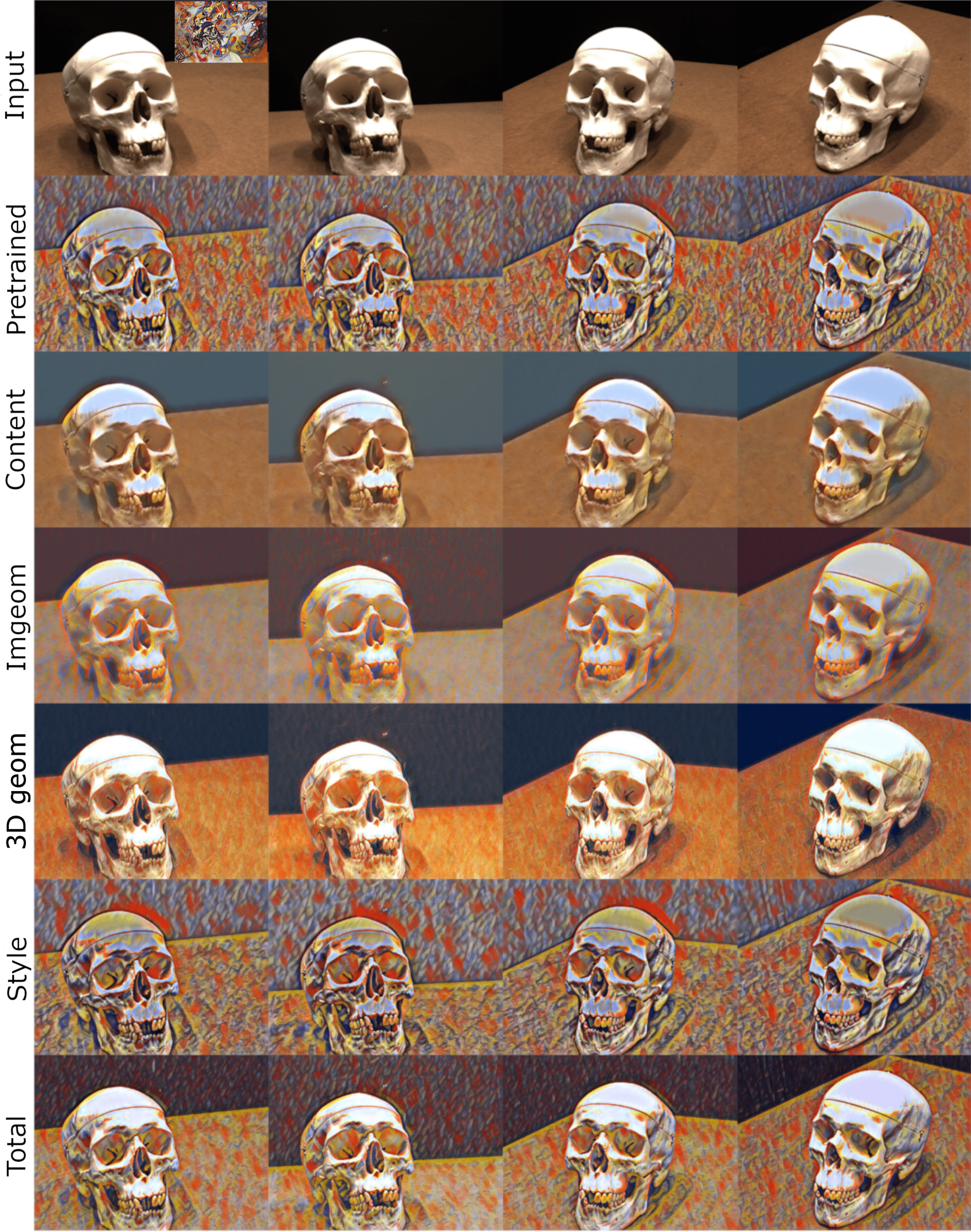}
	\caption{Ablation study results with PatchMatchNet\_UNet by separately training the network with different loss terms. The first row shows the RGB input, and the second row shows the results obtained from the pretrained network. The subsequent rows demonstrate the results of the network trained with each loss term separately, starting with the content loss in the third row, followed by image geometry, 3D geometry loss, and style loss in the fourth, fifth, and sixth rows, respectively. The last row demonstrates the results obtained by incorporating all the loss terms.}
	\label{fig:ablation_unet}
\end{figure*}

\begin{figure*}[h!]
	\centering
	\includegraphics[width=0.90\textwidth]{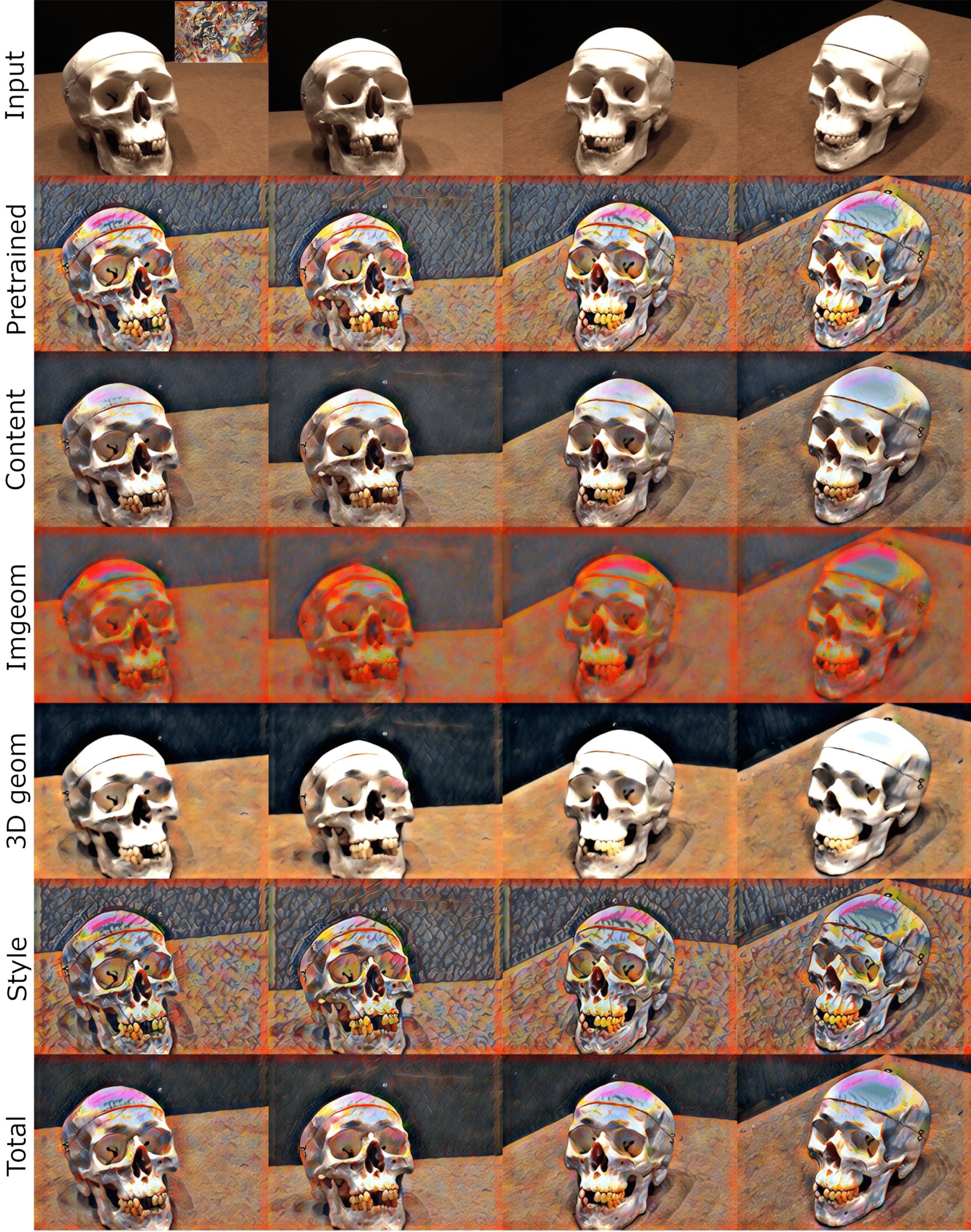}
	\caption{Ablation study results with PatchMatchNet\_AdaIN by separately training the network with different loss terms. The first row shows the RGB input, and the second row shows the results obtained from the pretrained network. The subsequent rows demonstrate the results of the network trained with each loss term separately, starting with the content loss in the third row, followed by image geometry, 3D geometry loss, and style loss in the fourth, fifth, and sixth rows, respectively. The last row demonstrates the results obtained by incorporating all the loss terms.}
	\label{fig:ablation_adain}
\end{figure*}

In this part, we aim to demonstrate the effectiveness and robustness of our pipeline with sparse views by conducting ablation experiments on loss terms.
To illustrate this, we utilize the 4 views of the scan65 dataset from DTU~\cite{aanaes2016_dtu} as an example and experiment with two different architectures, PatchmathNet\_UNet and PatchmathNet\_AdaIN, as shown in Figure \ref{fig:ablation_unet} and Figure \ref{fig:ablation_adain}, respectively. 

To examine the impact of each loss term, we train the network separately with each loss term and demonstrate the corresponding results. These results indicate that the UNet-based architecture recovers the original image features better with the help of content, image-geometry, and 3D geometry losses. On the other hand, the AdaIN-based architecture can recover the original image features using content and 3D geometry losses. However, the results of 3D geometry loss with AdaIN architecture tend to be blurrier than those of UNet. Additionally, we observed that the image-geometry loss pays attention to photometric edges, while the style loss tends to enhance artistic features in both cases.

From the experiments, we can conclude that the 3D geometry losses is effective in capturing the overall geometry of the scene. However, guiding the network with only this loss term may reduce the artistic features of the images.  Note that in our ablation studies, we combined volume and depth loss into a unified 3D geometry loss.
Specifically, we noticed that while volume loss struggles to converge independently, its integration alongside depth loss enhances pose estimation, particularly in scenarios involving low-resolution images. This integration offered promising results for improving the network's performance in handling varied image qualities and dimensions.

Meanwhile, the image geometry loss can be attentive to the photometric edges, and the content loss can preserve the original content of the images while reducing the artistic features. Moreover, the style loss can be useful in retaining artistic features. These findings motivated us to use a weighted sum of these loss terms to conduct geometry-aware style transfer.


\end{document}